\documentclass{article}

\usepackage{arxiv}

\usepackage[utf8]{inputenc} 
\usepackage[T1]{fontenc}    
\usepackage{hyperref}       
\usepackage{url}            
\usepackage{booktabs}       
\usepackage{amsfonts}       
\usepackage{nicefrac}       
\usepackage{microtype}      
\usepackage{lipsum}
\usepackage{graphicx}
\graphicspath{ {./images/} }
\usepackage{graphicx}
\usepackage{natbib}
\usepackage{doi}
\usepackage{amsmath}
\usepackage{caption}
\usepackage{subcaption}
\usepackage[linesnumbered,ruled,vlined]{algorithm2e}
\usepackage[mathscr]{euscript}
\usepackage{breqn}
\usepackage{amssymb}
\usepackage{latexsym}
\usepackage{cleveref}
\DeclareMathAlphabet      {\mathbfit}{OML}{cmm}{b}{it}
\title{A Novel Light Field Coding Scheme Based on Deep Belief Network \& Weighted Binary Images for Additive Layered Displays}

\author{\href{https://orcid.org/0000-0001-5262-6724}{\includegraphics[scale=0.06]{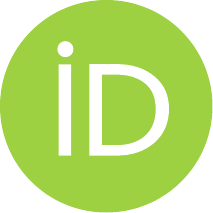}\hspace{1mm}Sally Khaidem}\\
	Department of Electrical Engineering, Indian Institute of Technology Madras, India \\
	\texttt{ee20d041@smail.iitm.ac.in} \\
	\\
	\href{https://orcid.org/0000-0003-3243-3321}{\includegraphics[scale=0.06]{orcid.pdf}\hspace{1mm}Mansi Sharma} \\
    Department of Computer Science and Engineering, Amrita School of Computing\\ Coimbatore, Amrita Vishwa Vidyapeetham, India\\
	Department of Electrical Engineering, Indian Institute of Technology Madras, India \\
	\texttt{s$\_$mansi@cb.amrita.edu, mansisharma@ee.iitm.ac.in} \\
}
\begin{document}
\maketitle
\begin{abstract}
Light-field displays create immersive experience by providing binocular depth sensation and motion parallax. Stacking light attenuating layers is one approach to implement a light field display with a broader depth of field, wide viewing angles and high resolution. Due to the transparent holographic optical element (HOE) layers, additive layered displays can be integrated into augmented reality (AR) wearables to overlay virtual objects onto the real world, creating a seamless mixed reality (XR) experience. This paper proposes a novel framework for light field representation and coding that utilizes Deep Belief Network (DBN) and weighted binary images suitable for additive layered displays. The weighted binary representation of layers makes the framework more flexible for adaptive bitrate encoding. The framework effectively captures intrinsic redundancies in the light field data, and thus provides a scalable solution for light field coding suitable for XR display applications. The latent code is encoded by H.265 codec generating a rate-scalable bit-stream. We achieve adaptive bitrate decoding by varying the number of weighted binary images and the H.265 quantization parameter, while maintaining an optimal reconstruction quality. The framework is tested on real and synthetic benchmark datasets, and the results validate the rate-scalable property of the proposed scheme.
\end{abstract}

\let\thefootnote\relax\footnotetext{The paper is under consideration at Pattern
Recognition  Letters.}

\section{Introduction}
Stereoscopic displays present a naturally immersive, intuitive visual interface for plenoptic contents with realistic disparity and smooth motion parallax~\cite{surman2014towards,li2020light,watanabe2019aktina}. They are generally categorised based on the necessity to wear specially designed glasses and the number of supported viewing angles. The most prevalent form of stereoscopic display needs passively polarised or rapidly alternating shuttered glasses to perceive the 3D effect. It provides depth perception by showing different images to the left and right eyes. Users typically detest donning invasive eyewear or attire that reduces their overall ambient visual acuity. Hence, there is a clear preference for non-invasive autostereoscopic displays that present conventional depth perception and natural motion parallax according to the viewer’s movements. 
\begin{figure}[!t]
     \centering
     \begin{subfigure}{0.4\textwidth}
        \centerline{\includegraphics[width=0.8\textwidth]{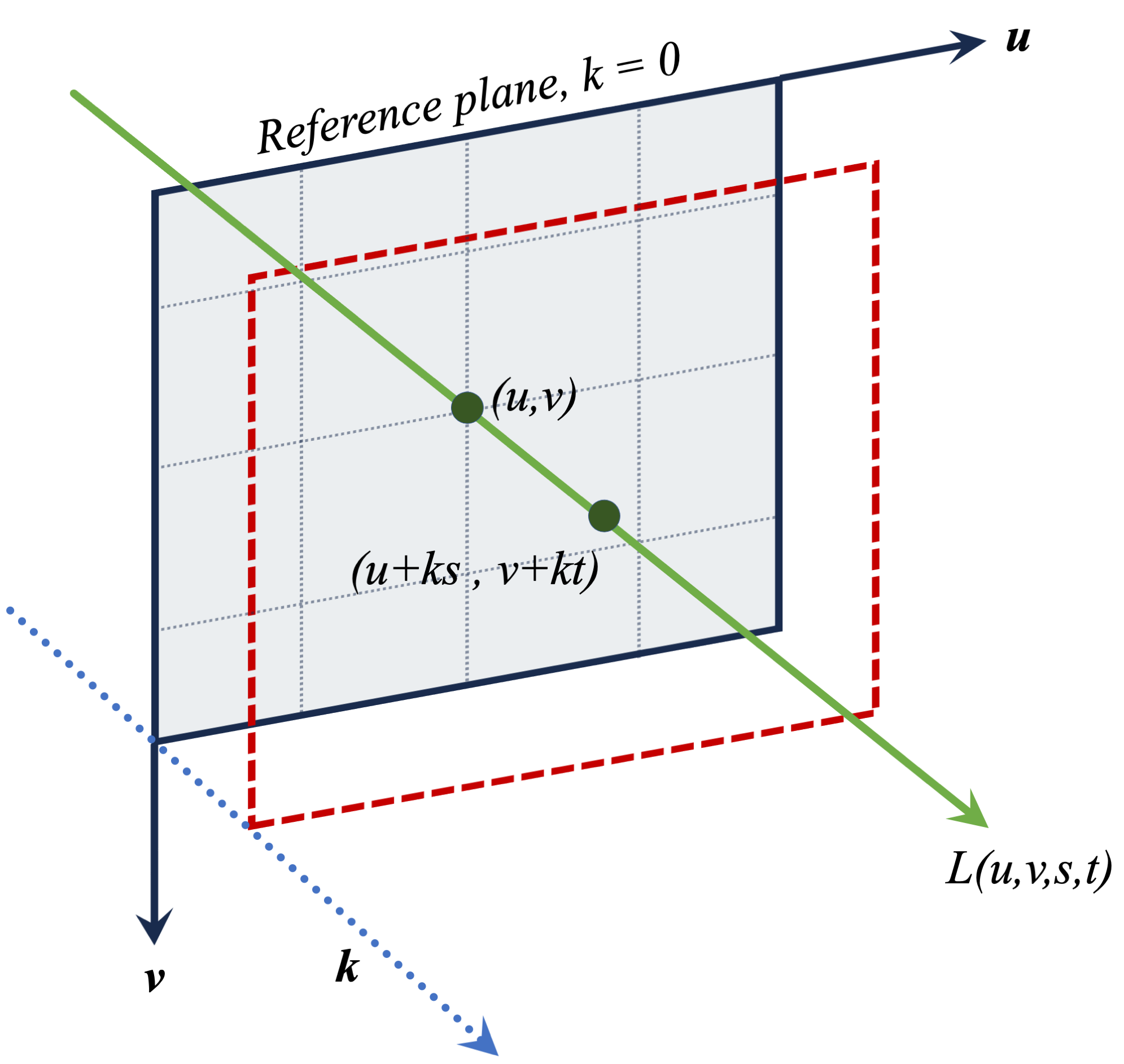}}
        \caption{}
        \label{lightfield}
     \end{subfigure}
     \begin{subfigure}{0.5\textwidth}
        \centerline{\includegraphics[width=0.9\textwidth]{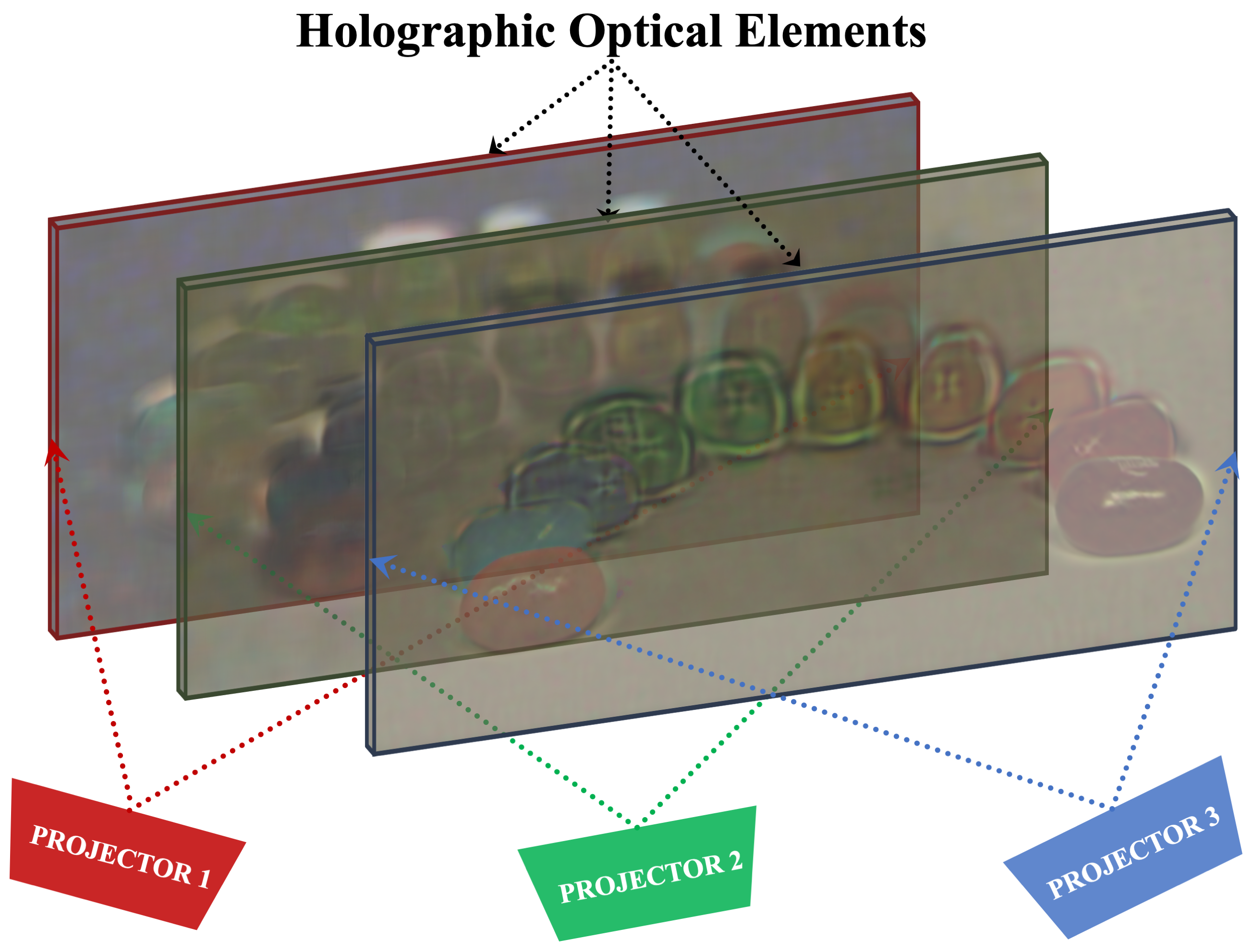}}
        \caption{}
        \label{layered_display}
     \end{subfigure}
     \caption{$(a)$ Simplified 4D light field, $(b)$ Structure of additive layered light-field display.}
\end{figure}

Researchers have employed various techniques to develop glasses-free multi-view/light-field displays. These methods include the use of parallax barriers~\cite{ives1903parallax, isono1993autostereoscopic, sakamoto2006multi, peterka2008advances}, specially designed lenses such as lenticular screens or integral photography lenses~\cite{lippmann1908epreuves, borner1993autostereoscopic, mccormick1995integral}, stacked layers~\cite{wetzstein2012tensor, takahashi2018focal} and layered display~\cite{maruyama2020comparison, 10.1145/2897824.2925971}. Extended depth of field, wider field of view, and thin form factor are desirable characteristics defining an excellent light field display. Therefore, incorporating these features is necessary to ensure the display produces a clear, accurate light field that provides users with a realistic and engaging viewing experience. 

Augmented reality (AR) technology allows digital information to be overlaid in the real world, creating a composite view. Flat displays in current AR wearables generate conflicting depth information. It lacks monocular depth cues, resulting in three significant visual conflicts: vergence accommodation conflict, focal rivalry and ocular parallax \cite{kramida2015resolving, konrad2020gaze}. The light field display can be integrated into AR wearables by overlaying 3D digital information onto the real world without visual conflicts~\cite{sluka2021digital}. 

We illustrate the conventional structure of an additive layered display in Fig.~\ref{layered_display}. This display uses holographic optical elements (HOEs) as transparent layers to diffuse projected images from projectors, creating separate 2D light fields. The combined light rays from each layer pass through different pixel combinations based on viewing angles, resulting in a 4D light field. Additive light field displays do not suffer from the moiré effect, a visual artefact that decreases brightness and colour accuracy in LCD-based compressive displays. The transparent holographic optical elements (HOEs) make additive layered displays well-suited for augmented reality (AR) applications. This technology offers users a highly immersive and interactive way of experiencing digital content, allowing them to engage with virtual objects more naturally and intuitively.

Plenoptic data contains detailed information about the direction and intensity of light rays in a scene, resulting in a large amount of data with spatial, angular, and temporal redundancies. It is essential to consider these inherent correlations to create an effective compressive content delivery pipeline via multi-layered based tensor displays. Many currently available methods for encoding light field data are unsuitable for multi-layered display technologies. Various encoding algorithms extract sub-aperture images (SAIs) and create a pseudo video sequence in light field data compression~\cite{liu2016pseudo,li2017pseudo}. Commonly used video encoders like HEVC~\cite{sullivan2012overview} or MV-HEVC~\cite{hannuksela2015overview} are employed for inter and intra-frame hybrid prediction. However, current view estimation-based coding methods cannot remove redundancies among adjacent SAIs and are limited to local or frame units of the encoder~\cite{senoh2018efficient,huang2018view,heriard2019light}. Learning-based view-synthesis techniques require a vast and diverse dataset to achieve better compression~\cite{bakir2018light,wang2019region,jia2018light,liu2021view}. Some techniques use the low-rank structure in light field data based on disparity models, while others use light field structure/geometry to compress at low bitrates~\cite{vagharshakyan2017light,ahmad2020shearlet,chen2020light}. However, these methods are still are not suitable for layer pattern encoding for additive layered displays.

This paper introduces a novel method for encoding layer patterns of additive light field displays in a scalable and efficient manner. The proposed scheme involves using a convolutional neural network (CNN) to obtain the optimised layer pattern for the display. Our aim of achieving a scalable framework led us to convert the layers into their weighted binary image form. The binary images and their appropriate weight can reconstruct the layers with minimal distortion, and the reconstruction accuracy improves as more binary images are considered. Moreover, the divide-and-conquer strategy inherent in the scalable structure significantly reduces computational complexity. The weighted binary image is highly compressible, as the pixel values take either $0$ or $1$ to represent black and white pixels, respectively. We employ a powerful generative model to exploit the inherent strong correlations in binary images. The deep autoencoder model learns low dimensional codes that reduce the dimensionality of data. Our deep architecture uses multiple stacks of Restricted Boltzmann Machines (RBM)~\cite{teh2000rate}, which form the Deep Belief Network (DBN)~\cite{hinton2006reducing}. In DBN, the hidden layer (learned features) of an RBM feeds to the visible layer on the next RBM on the stack. Finally, the widely adopted HEVC (HM 17.0)~\cite{sullivan2012overview} standard video codec is employed on the latent code to compress further and eliminate intrinsic redundancies in latent data blocks, generating a bitstream compatible with most decoder devices. Our tests with both real-world and synthetic light field data demonstrate highly competitive outcomes. The principal contributions introduced in the proposed light field encoding technique can be summarized as:

\begin{itemize}
    \item{The paper presents a new light field coding method for additive layered displays, which effectively eliminates spatial, temporal, angular, and non-linear redundancies between adjacent sub-aperture images in a single integrated framework. The first component of the proposed approach operates to eliminate both intra-view and inter-view redundancies, leading to the derivation of distinct layer patterns. The second block effectively mitigates highly correlated non-linear redundancies among the patterns. Overall, the proposed framework represents a highly efficient and comprehensive solution for light field coding.}
    \item{The integrated and versatile framework of weighted binary images and deep belief networks (DBN) facilitates scalability in light field reconstruction. The number of levels utilized in the proposed approach plays a critical role in determining the reconstruction quality, whereby progressively increasing levels correspond to improved accuracy. The proposed framework thus represents a powerful solution for high-quality reconstruction with support for multiple bitrates, all within a single integrated pipeline.}
    \item{In the proposed scheme, hybrid convolutional neural networks (CNN) and deep belief networks (DBN) offer adaptability across multiple bitrates, providing a highly flexible solution for light field coding. This scheme differs from traditional light field coding approaches that typically support fixed bitrate during compression. Furthermore, the proposed approach employs optimised layer patterns to process the light field data efficiently, eliminating the need to process the entire data set.}
\end{itemize}
\begin{figure*}[!t]
     \centering
     \begin{subfigure}{0.35\textwidth}
        \centerline{\includegraphics[width=0.9\textwidth]{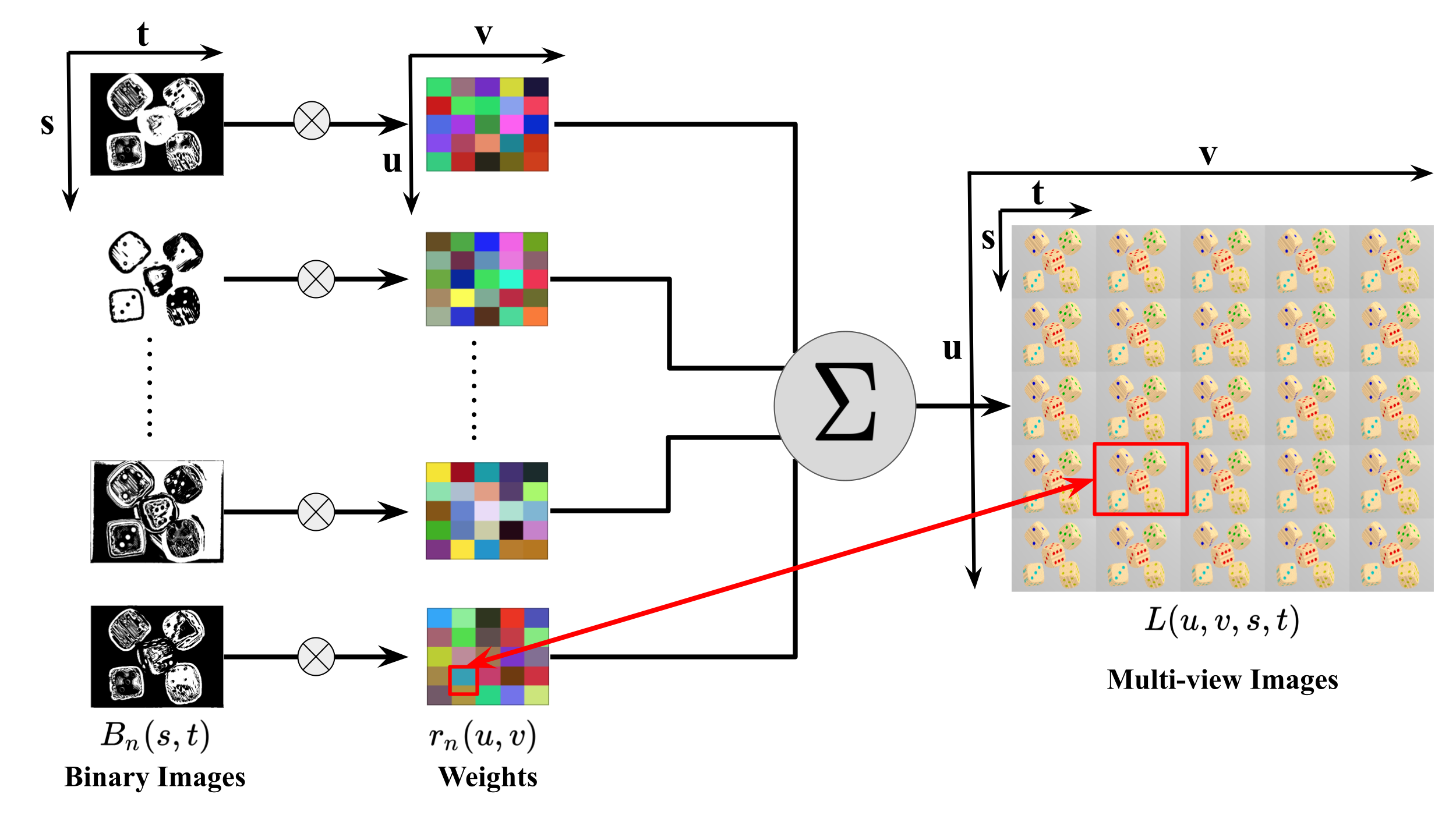}}
        \caption{}
        \label{binary_pipeline}
     \end{subfigure}
     \begin{subfigure}{0.3\textwidth}
        \centerline{\includegraphics[width=0.9\textwidth]{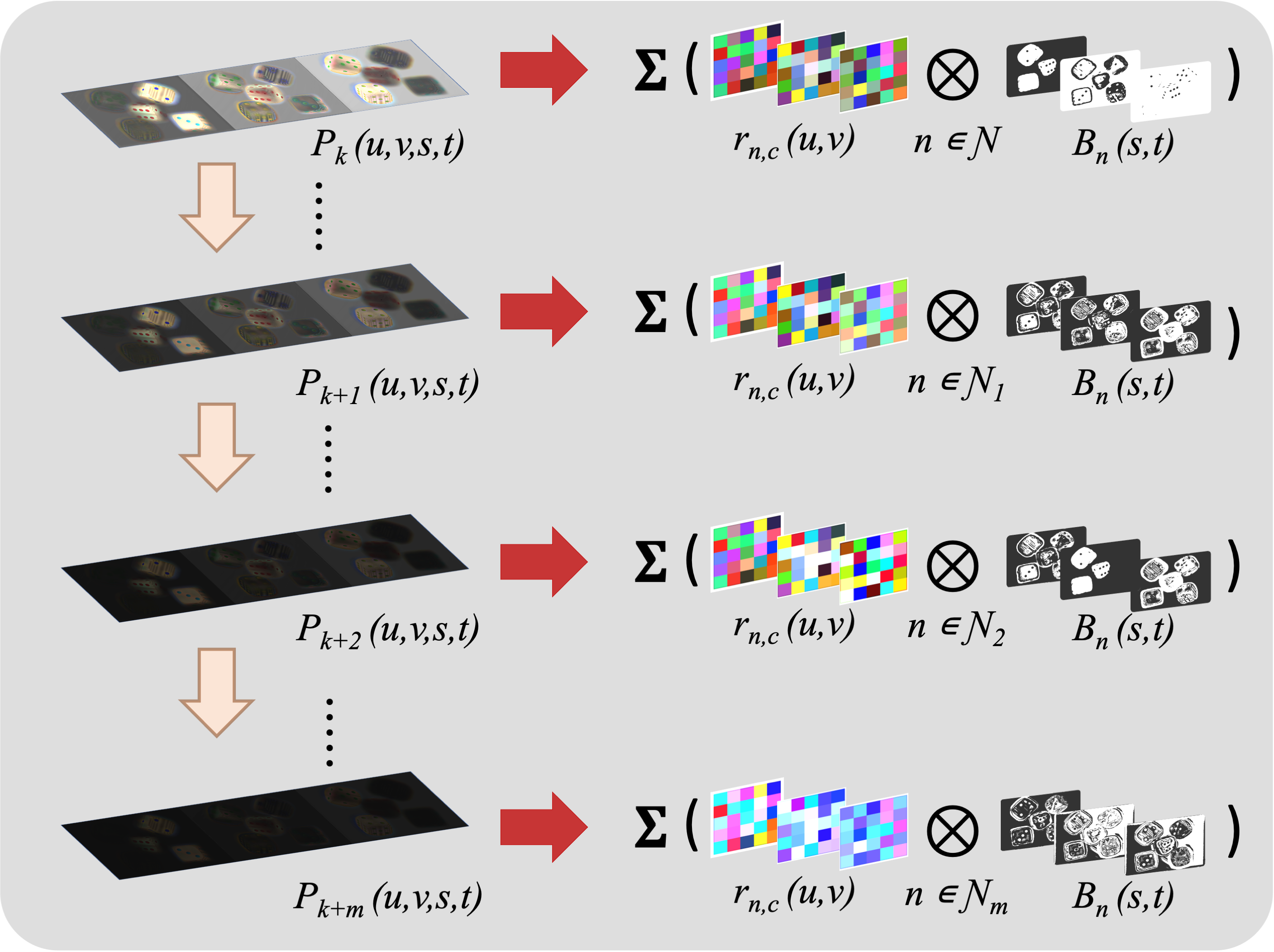}}
        \caption{}
        \label{scalablepipeline}
     \end{subfigure}
    \begin{subfigure}{0.33\textwidth}
        \centerline{\includegraphics[width=0.7\textwidth]{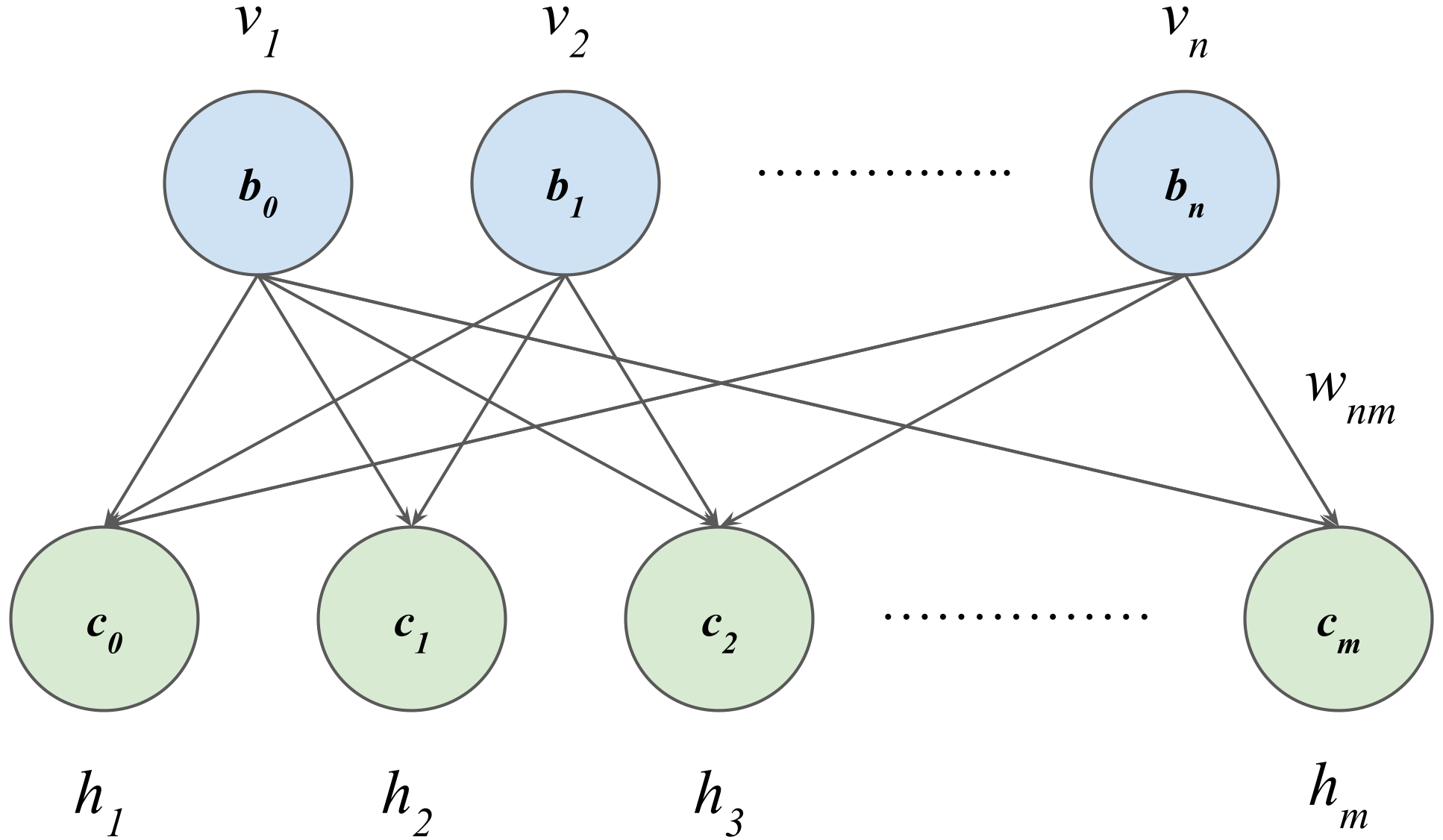}}
        \caption{}
        \label{rbm_structure}
     \end{subfigure}
     \caption{$(a)$ Compressed light field representation using binary images $B_n(s,t)$ and weights $r_n(u,v)$ $(b)$ Workflow of scalable encoding process $(c)$ Structure of restricted boltzmann machine network.}
\end{figure*}

\begin{figure*}[!t]
\centerline{\includegraphics[width=0.8\textwidth]{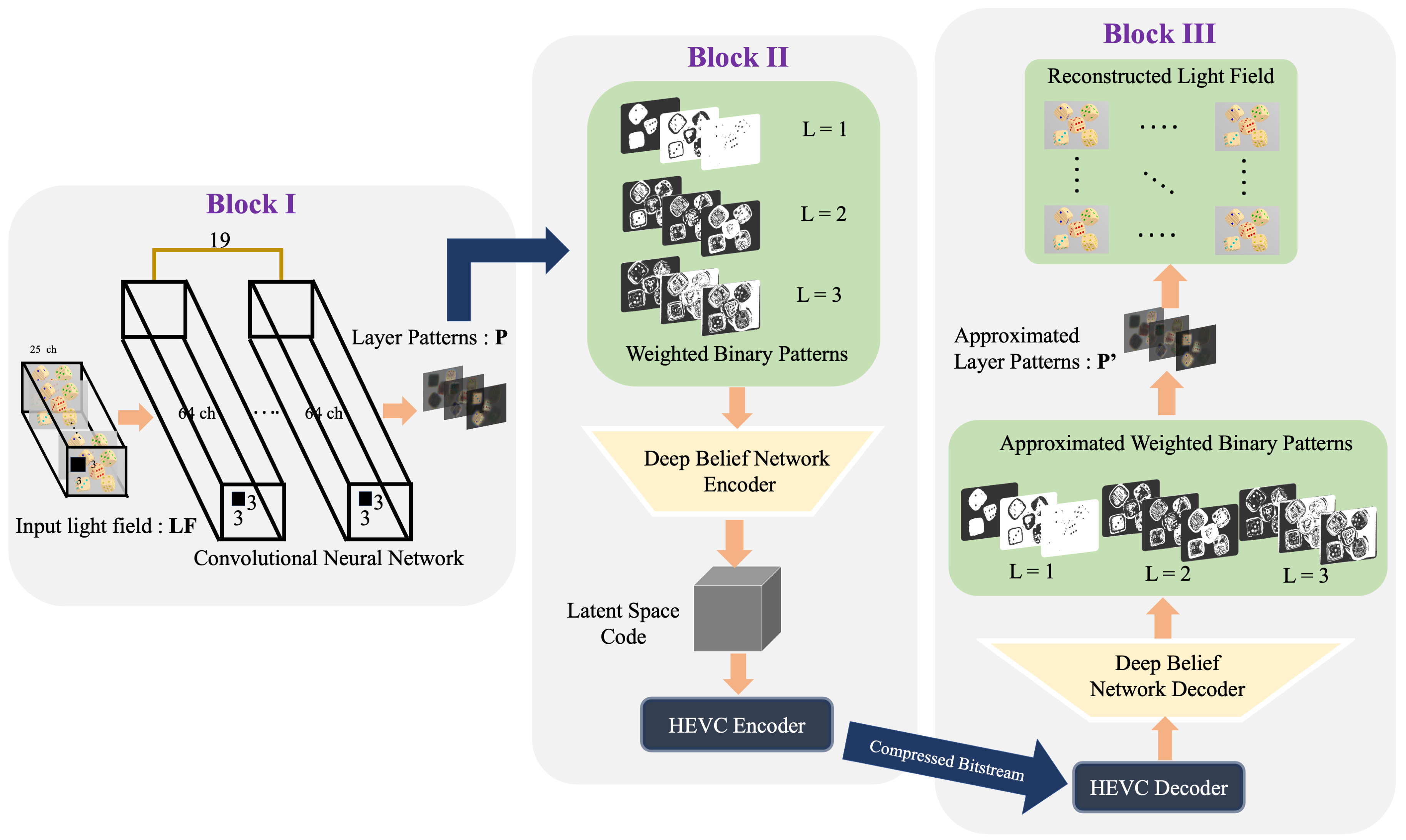}}
\caption{Complete workflow of proposed coding scheme.}
\label{Pipeline}
\end{figure*} 

\section{Proposed Coding Scheme for Multi-Layered Displays}

Our representation and coding pipeline consists of three blocks. The complete workflow is illustrated in Fig.~\ref{Pipeline}. BLOCK 1 (Fig.~\ref{Pipeline}) transforms the entire light field into additive layer patterns using a CNN-based approach. A more efficient and scalable representation of the layer pattern is achieved through a DBN and HEVC encoding in BLOCK II. The light field reconstruction from approximated layers is performed in BLOCK III. The implementation steps of the proposed methodology is described in Algorithm ~\ref{alg:pseudocode1}.

The multidimensional plenoptic function represents the amount of light travelling through each point in space in each direction. We adopt an angle + plane parameterization with a reference plane at $k=0$ as shown in Fig.~\ref{lightfield}. The 7-D function,  $L(x,y,z,\theta,\phi,\lambda,t)$ is constructed by measuring light ray at every potential incidence $(\theta,\phi)$, wavelength $\lambda$, and time $t$. We parameterised light ray by the point of intersection with the reference plane $[(u,v)]$ and the outgoing direction with respect to the z-axis $[(\theta,\phi)]$. Hence, it is simplified into a 4-D function $L(u,v,s,t)$, where $s = \tan(\theta)$ and $t = \tan(\phi)$. In the light field model, $st$ is considered the plane for a set of cameras having $uv$ as their focal plane.  Hence, $st$ and $uv$ are called the angular and spatial resolution, respectively.
\subsection{Additive Layer Pattern Generation}
The main objective of an autostereoscopic 3D display is to support simultaneous viewing from multiple perspectives without compromising the resolution of each viewpoint. An example of such transparent tensor display is the ``additive layered display'' illustrated in Fig.~\ref{layered_display}. The display consists of transparent HOE projection layers which diffuse images from projectors generating independent 2D light fields. The resulting 2D light fields are merged by addition operation into a 4D light field, providing motion parallax according to the viewing position. The emitting light ray is formulated as shown in Equation~\ref{eq:1}.
\begin{equation}
\begin{aligned}
    L_{add}(u,v,s,t) = \sum_{k\in Z}P_k(u+ks,v+kt)
    \label{eq:1}
\end{aligned}
\end{equation}
where, $P_k(u, v)$ denotes transmittance of the layer located at $k \in Z$. We assumed that there are three layers in the light field display located at $k = {-1, 0, 1}$.

A CNN-based network optimises additive layers to display the target 3-D scene as shown in \textbf{Block I} of Fig.~\ref{Pipeline}. The network consists of sequentially stacked 20 2D convolutional layers. The spatial size of the tensors is constant, while only the number of channels is varied throughout the network. The tensors $\textbf{L}$ and $\mathbf{L_{add}}$ have 25 channels each corresponding to the $5\times 5$ viewpoints. While tensor \textbf{P} has three channels each for the layers in the display, the intermediate feature maps have 64 channels. The mapping function of the optimisation process is expressed as

\begin{equation}
\begin{aligned}
    f:L\to P
    \label{eq:2}
\end{aligned}
\end{equation}
where, $L$ denotes the tensor with all pixels of $L(u,v,s,t)$ for all $(u,v)$ and $P$ represents the tensor with all pixels of $P_k(s,t)$ for all $k\in Z$. The mapping from layer patterns back to light field can be expressed as

\begin{equation}
\begin{aligned}
    g:P\to L_{add}
    \label{eq:3}
\end{aligned}
\end{equation}
where, $L_{add}$ denotes all light rays in $L_{add}(u,v,s,t)$. The CNN is constructed such that it corresponds to the composite mapping of $g \circ f$ minimising squared loss error as expressed in ~\eqref{eq:4}.
\begin{equation}
\begin{aligned}
   \underset{f}{argmin} = \left\|L - L_{add}\right\|^2
    \label{eq:4}
\end{aligned}
\end{equation}
\subsection{Scalable Coding with Weighted Binary Images}
The obtained layer patterns are transformed into a weighted binary representation, where binary images with defined weights can approximate the layers~\cite{8451812}. There are only two possible intensity values (black and white) in binary images. They are notably helpful for segmentation and thresholding applications in image processing. The compressed representation (Fig.~\ref{binary_pipeline}) of layer patterns is possible using weighted binary images. The binary images $B_n$ and their corresponding weights of `$c$' channel, $r_{n,c}$ can optimally approximate the layers pattern $P_k$ as formulated in ~\eqref{eq:binary}. The binary image captures common features for all viewpoints and the pixel-independent weights contain the differences between the layers. 

\begin{equation}
\begin{aligned}
    P_k(u,v,s,t) \simeq  \sum_{n=1}^{N}B_n(s,t)r_{n,c}(u,v)
    \label{eq:binary}
\end{aligned}
\end{equation}

\begin{figure}[!t]
     \centering
     \begin{subfigure}{0.122\textwidth}
        \centerline{\includegraphics[width=1\textwidth]{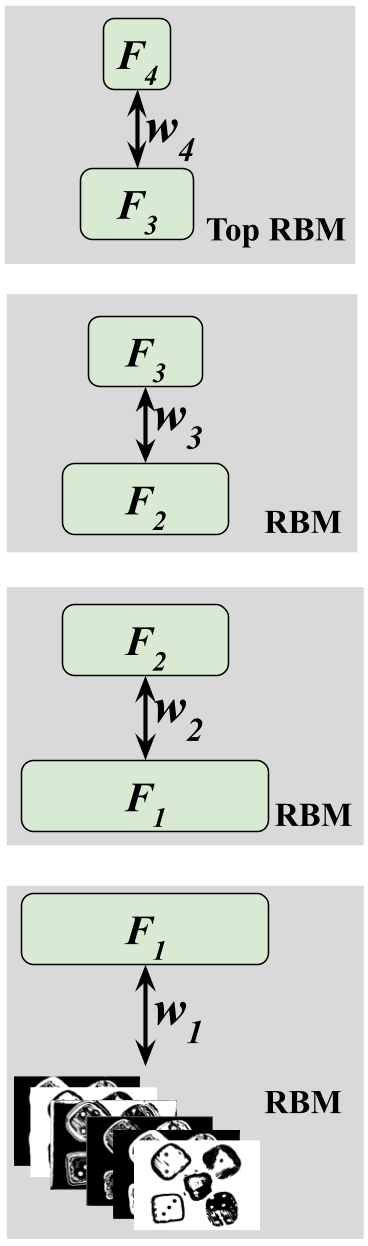}}
        \caption{}
        \label{pretraining}
     \end{subfigure}
     \begin{subfigure}{0.13\textwidth}
        \centerline{\includegraphics[width=1\textwidth]{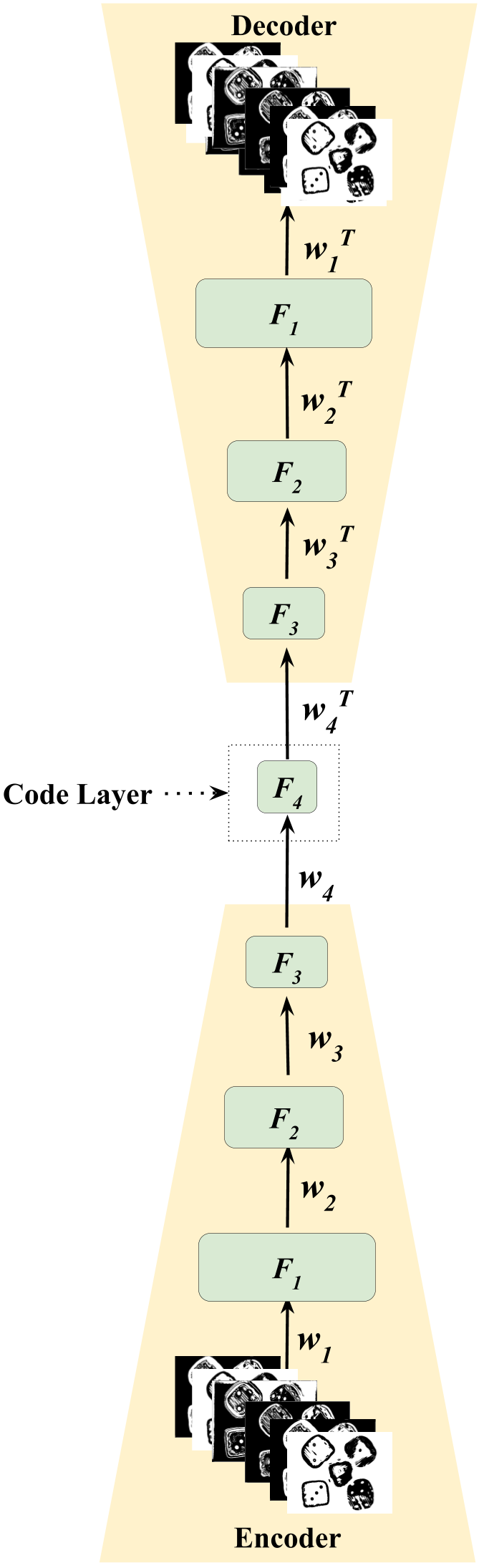}}
        \caption{}
        \label{unrolling}
     \end{subfigure}
     \begin{subfigure}{0.13\textwidth}
        \centerline{\includegraphics[width=1\textwidth]{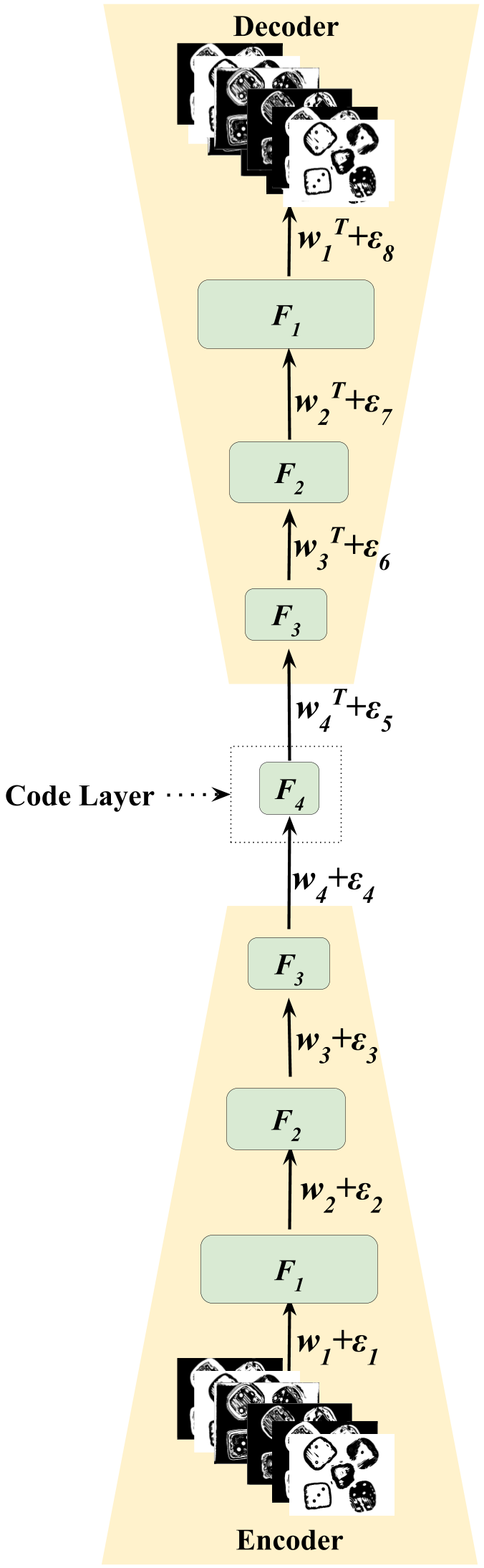}}
        \caption{}
        \label{finetuning}
     \end{subfigure}
        \caption{Workflow of deep auto-encoder with RBM pretraining.$(a)$ Pre-training of RBM weights $(b)$ Unrolling of stacked RBMs $(c)$ Fine-tuning of the unrolled stacked RBMs network.}
        \label{DBN}
\end{figure}
In order to obtain the compressed binary representation, it is necessary to solve the optimisation problem defined in ~\eqref{eq:binaryoptimisation} 
\begin{equation}
\begin{aligned}
\begin{split}
    \underset{B_n(u,v),r_{n,c}(u,v)}{argmin}\sum_{u,v,s,t}\left| P_k(u,v,s,t)-\sum_{n=1}^{N}B_n(s,t)r_{n,c}(u,v)\right|^2
\end{split}
\label{eq:binaryoptimisation}
\end{aligned}
\end{equation}
The pipeline adopts two alternate approaches for the two unknowns, $B_n$ and $r_{n,c}$. It initialises the binary images $B_n$ and repeats the following two steps until convergence. 
\begin{enumerate}
    \item Optimise the weights $r_{n,c}(u,v)$ keeping binary images $B_n(s,t)$ fixed. This step involves solving standard least squares problems.
    \item Optimise binary images $B_n(s,t)$ keeping the weights $r_{n,c}(u,v)$ fixed. The individual pixels $(s,t)$ are solved using binary combinatorial optimisation technique which is a NP-Hard problem~\cite{toth2000optimization}.
\end{enumerate}
Since combinatorial optimisation employs exhaustive search, the introduction of scalability in the framework plays a vital role in drastically increasing the speed and reducing computational complexity.
For scalable coding (Fig.~\ref{scalablepipeline}), the divide and conquer method is adopted. The $N$ number of layer patterns are divided into $M$ groups, which correspond to the number of levels for scalability. Firstly, $M$ sets of integers are defined such that $\left\{N_1,...,N_m\right\}$ satisfy $N_i \cap N_j =\phi$ for $i\neq j$ and $\bigcup _{m=1...M}\ N_m=\left\{1,..,N\right\}$. $N_m$ denotes the binary images for the $m^{th}$ level and $\sum_{m}\left| N_m \right|=N$. For original $N$ bit binary combinatorial optimisation, the computational complexity is $\mathcal{O}(2^N)$. The complexity is reduced to $\mathcal{O}(\sum_m 2^{\left | \mathscr{N}_m \right |})$ when divided into $M$ sublevels. The approximation accuracy of the encoding process improves progressively as the number of levels increases. The target layer patterns $P_{k+1}(u,v,s,t)$ is set to original layer pattern $P_{k}(u,v,s,t)$ in the first level. The best approximation of the layer patterns is evaluated at each level using the binary images $B_n(s,t)$ along with their corresponding weights $r_{n,c}(u,v)$ and are optimised as shown in \eqref{eq:binaryoptimisationscalable}.
\begin{equation}
\begin{aligned}
\begin{split}
    \underset{B_n(u,v),r_{n,c}(u,v)}{argmin}\sum_{u,v,s,t}\left| P_{k+M}(u,v,s,t)-\sum_{n\in N_M}B_n(s,t)r_{n,c}(u,v)\right|^2
\end{split}
\label{eq:binaryoptimisationscalable}
\end{aligned}
\end{equation}

The calculation of target layer pattern for the next level using residual from the first layer is shown in \eqref{eq:residualscalable}
\begin{equation}
\begin{aligned}
\begin{split}
    P_{k+1}(u,v,s,t)=P_{k}(u,v,s,t)-\sum_{n\in N_M}B_n(s,t)r_{n,c}(u,v)
\end{split}
\label{eq:residualscalable}
\end{aligned}
\end{equation}
The steps in \eqref{eq:binaryoptimisationscalable} and \eqref{eq:residualscalable} are repeated to the $M^{th}$ level. Hence, a scalable representations with $M$ levels is realised as
\begin{equation}
\begin{aligned}
    P_k(u,v,s,t) \simeq  \sum_{m=1}^{M}\sum_{n\in N_m}B_n(s,t)r_{n,c}(u,v)
    \label{eq:binary_colour}
\end{aligned}
\end{equation}
As the level progresses, the residual \eqref{eq:residual} exhibits a monotonic decrease, indicating that less information is available for the higher level. This decrease is accompanied by a progressive improvement in the accuracy of the decoded layer patterns, which is beneficial in various real-world scenarios, such as adaptive rate control and flexible user adaptability. The progressive improvement in accuracy is a direct result of the increasing number of levels, which allows for the incorporation of finer details in the reconstructed image. These findings suggest that the proposed method is well-suited for a wide range of applications that require high-quality reconstructed images with adaptive resolution.
\begin{equation}
\begin{aligned}
    \left|P_{k+1}(u,v,s,t) \right|\leq \left|P_{k}(u,v,s,t) \right|
    \label{eq:residual}
\end{aligned}
\end{equation}

\subsection{Deep Belief Network}
\begin{algorithm}[t!]
\setcounter{AlgoLine}{0}
\DontPrintSemicolon
\SetKwInput{KwInput}{Input}                
\SetKwInput{KwOutput}{Output}              
\KwInput{Input light field $\mathbfit{L(u,v,s,t)}$}
\KwOutput{Non-linear low dimensional code, $\mathbfit{NLD}$}

Initialize $\mathbfit{P_k(u+ks,v+kt)}$ where $u$ and $v$ are limited to the range $[-2,2]$ and $Z = \{-1, \ 0, \ +1\}$ \label{alg:l1}

\Repeat{optimisation problem is solved}
{
      \For{$  k \in Z $}
       {
         $\mathit{arg \ \underset{P_k|k\in Z }{min}} \sum\limits_{u,v,s,t}\left \| L(u,v,s,t) - \sum\limits_{k \in Z} P_k(u+ks,v+kt)) \right \|^2 $  \label{alg:l2}
       }
}

Initialise first level ($m=1$) of binary weighted pattern $\mathbfit{W_1(u,v,s,t) \Leftarrow P_k(u,v,s,t)}$

\For{ $m = 1 \ to \ M $}
{
Obtain $\mathbfit{B_n(s,t)}$,  $\mathbfit{r_{n,c}(u,v)}$ where $n\in \mathscr{N}_m$ by solving
{
$\mathit{arg \ \underset{B_n(s,t) r_{n,c}(u,v)}{min}} \sum\limits_{u,v,s,t}\left \| P_k(u,v,s,t) - \sum\limits_{n\in \mathscr{N}_m} B_n(s,t)r_{n,c}(u,v) \right \|^2 $ \label{alg:l3}
}\\
Carry over the \textit{residual} as\\
{
${W_{m+1}(u,v,s,t)} = {W_{m}(u,v,s,t)} - {\sum\limits_{n\in \mathscr{N}_m}B_n(s,t)r_{n,c}(u,v)}$
}\\
$m \Leftarrow m+1$
}
 Dimensionality reduction of $\mathbfit{B_n(s,t)}$ for $\mathbfit{n \in \mathscr{N}_m}$\\
\For{n = 1 to $\mathscr{N}_m$}
{
$\mathbfit{NLD} \leftarrow \mathbfit{Algorithm \ 2}\ (\mathbfit{B_n(s,t)}) $
}
\caption{Proposed scalable encoding coding scheme}
\label{alg:pseudocode1}
\end{algorithm}
The binary layer patterns $B_n(s,t)$ and their weights $r_{n,c}(u,v)$ provides an approximate representation of the actual layer patterns. Algorithms such as auto-encoder encode the input data into a much smaller dimensional representation and store latent information about the input data distribution. An RBM is a two-layered stochastic network with visible and hidden layers(Fig.~\ref{rbm_structure}). It is a probabilistic model composed of weights and biases. The structure consists of $``n"$ visible units $\mathbf{v=(v_1,..,v_n)}$ representing the observed data and $``m"$ hidden units $\mathbf{h=(h_1,..,h_m)}$ to illustrate the dependencies in the observed data. There is no interconnection among the nodes in each layer to ensure their mutual independence. In binary RBMs, the random variables take the value $(v,h)\in \left\{0,1\right\}^{m+n}$. The hidden and visible variable vectors, $\mathbf{v}$ and $\mathbf{h}$ can be represented by their joint probability density as
\begin{equation}
\begin{aligned}
    p(\mathbf{v},\mathbf{h}) = e^{-E(\mathbf{v},\mathbf{h})}\bigg/\int \int_{\mathbf{v},\mathbf{h}}e^{-E(\mathbf{v},\mathbf{h})}
    \label{eq:probability_density_rbm}
\end{aligned}
\end{equation}
where $E(\mathbf{v},\mathbf{h})$ is the associated energy function. Since the input data is binary-value, we apply binary-binary energy function~\cite{shang2014data} as described in~\eqref{eq:energy_rbm}.
\begin{equation}
\begin{aligned}
    E(\mathbf{v},\mathbf{h})=-\sum_{i=1}^{m}\sum_{j=1}^{n}w_{ij}h_iv_j-\sum_{j=1}^{n}b_jv_j-\sum_{i=1}^{m}c_ih_i
    \label{eq:energy_rbm}
\end{aligned}
\end{equation}
where, $w_{ij}$ is the weight associated with the nodes $v_i$ and $h_j$. The bias terms for $j^{th}$ visible and $i^{th}$ hidden node are $b_j$ and $c_i$  respectively. The network assigns a probability to every possible image using the energy function. Adjusting the weights and biases to reduce the energy of a training image, increases the probability of that image. The weights are adjusted as shown in~\eqref{eq:weights}.
\begin{equation}
\begin{aligned}
    \Delta w_{ij} = \varepsilon (\left \langle v_ih_j \right \rangle_{data}-\left \langle v_ih_j \right \rangle_{recon}) 
    \label{eq:weights}
\end{aligned}
\end{equation}
where $\varepsilon$ is the learning rate, $\left \langle v_ih_j \right \rangle_{data}$ and $\left \langle v_ih_j \right \rangle_{recon}$ are the fraction of times pixel $i$ and feature $j$ are on together when driven by data and reconstruction data respectively.
\begin{algorithm}[t!]
\setcounter{AlgoLine}{0}
\DontPrintSemicolon
\SetKwInput{KwInput}{Input}                
\SetKwInput{KwOutput}{Output}              
\KwInput{$\mathbfit{DBN}$, training set $\mathbfit{X}$}
\KwOutput{Pre-trained $\mathbfit{DBN}$}
Initialize network parameters $\mathbfit{w,b}$\\
$\mathbfit{Input} \leftarrow \mathbfit{X}$\\
\For{ all RBM in DBN}
{
\For{ \textit{epoch} = 1 to e} 
{
\For{ k = 1 to floor($N_{sample}/N_{batch \ size}$)}
{
$\mathbfit{B} \leftarrow \ batch  \ from \ \mathbfit{Input}$

$\mathbfit{\Delta{w},\Delta{b}} \leftarrow k\ Contrastive\ Divergence$ 

$\mathbfit{w \leftarrow w + r.\Delta{w}}$\\
$\mathbfit{b \leftarrow b + r.\Delta{b}}$\\
}
}
$\mathbfit{X \leftarrow Input \times w + b}$
}
\caption{Layer-by-layer pre-training of DBN}
\label{alg:pseudocode2}
\end{algorithm}
A single layer of binary RBM is not the most effective technique to model the input data structure. A deep belief network (DBN) is a stack of RBMs (Fig.~\ref{DBN}) where the hidden layer of one RBM is the visible layer of next RBM. Stacking individual RBMs to form a chain drastically improves the performance, similar to a multi-layer perceptron outperforming a single perceptron. In DBN, the training procedure happens in two stages: unsupervised layer-by-layer pre-training and fine-tuning. During pre-training, we employ a layer-wise greedy technique(Fig.~\ref{pretraining}). Once the training of previous RBM completes, its hidden layer is the visible layer for the next RBM. Likewise, the training of all RBMs occur one after another until the last RBM is trained. After pretraining completion, the model unfolds to form a deep auto-encoder network initialised with the pre-trained weights (Fig.~\ref{unrolling}). The training of each RBM maximises the probability of its input data exploiting contrastive divergence (CD) \cite{hinton2002training} algorithm to update the network parameters. Each feature layer detects the strong and high-order correlations between the units in the layer beneath it. The pre-trained weights are fine-tuned (Fig.~\ref{finetuning}), minimising the cost function. We implement the back-propagation algorithm to update the whole network's parameters, progressively passing the error from the last layer to the bottom input layer. The pipeline achieves latent space representation of the weighted binary data using a DBN. The number of features extractors $F_1$, $F_2$, $F_3$ and $F_4$ in each layers are varied according to the requirement of input data such that $ F_2 > F_3 > F_4$ and $F_2 > F_1$ . The code layer with $F_4$ features is the latent space representation of the input data. The encoder and decoder network share a symmetrical structure and number of features in each layer. The latent code are encoded using standard video codec HEVC (HM 17.0). The implementation details and experimental results are analysed in the later sections.

\section{Experiments}
To evaluate the performance of the proposed scheme, we perform experiments on both real and synthetic light fields. We choose $Amethyst$, $Bunny$, and $Jelly Beans$ light fields from the Stanford Light Field Archive~\cite{vaish2008new} as well as the $Boxes$~\cite{honauer2017dataset}, $Dice$, $Dragon$ and $Bunnies$~\cite{marwah2013compressive} synthetic light fields. As illustrated in BLOCK I of Fig.~\ref{Pipeline}, we take the complete light field data as input to generate an optimized additive layer patterns. The additive layer patterns are represented as scalable weighted binary maps, which are compressed using Deep Belief Networks in BLOCK II with variable bitrate support. The compressed data is further encoded using HEVC to generate an encoded bitstream, which was subsequently decoded and reconstructed back into the complete light field in BLOCK III. 

\subsection{Implementation Details}
\begin{figure}[!h]
\renewcommand{\figurename}{Figure S}
     \centering
     \begin{subfigure}[b]{1\textwidth}
         \centering
         \includegraphics[width=.65\textwidth, height=0.155\textheight]{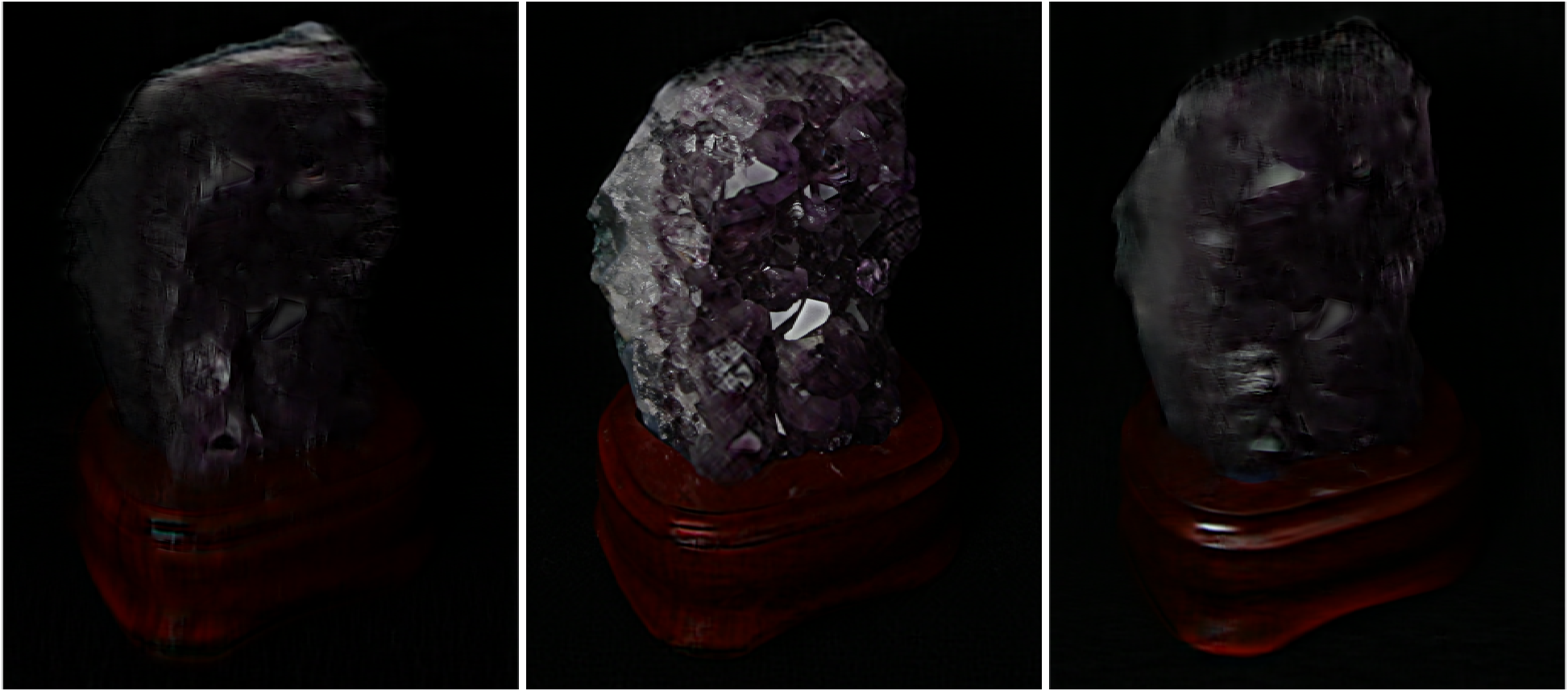}
     \subcaption{Original Layer Pattern}  
     \end{subfigure}
    \begin{subfigure}[b]{1\textwidth}
         \centering
         \includegraphics[width=.65\textwidth, height=0.155\textheight]{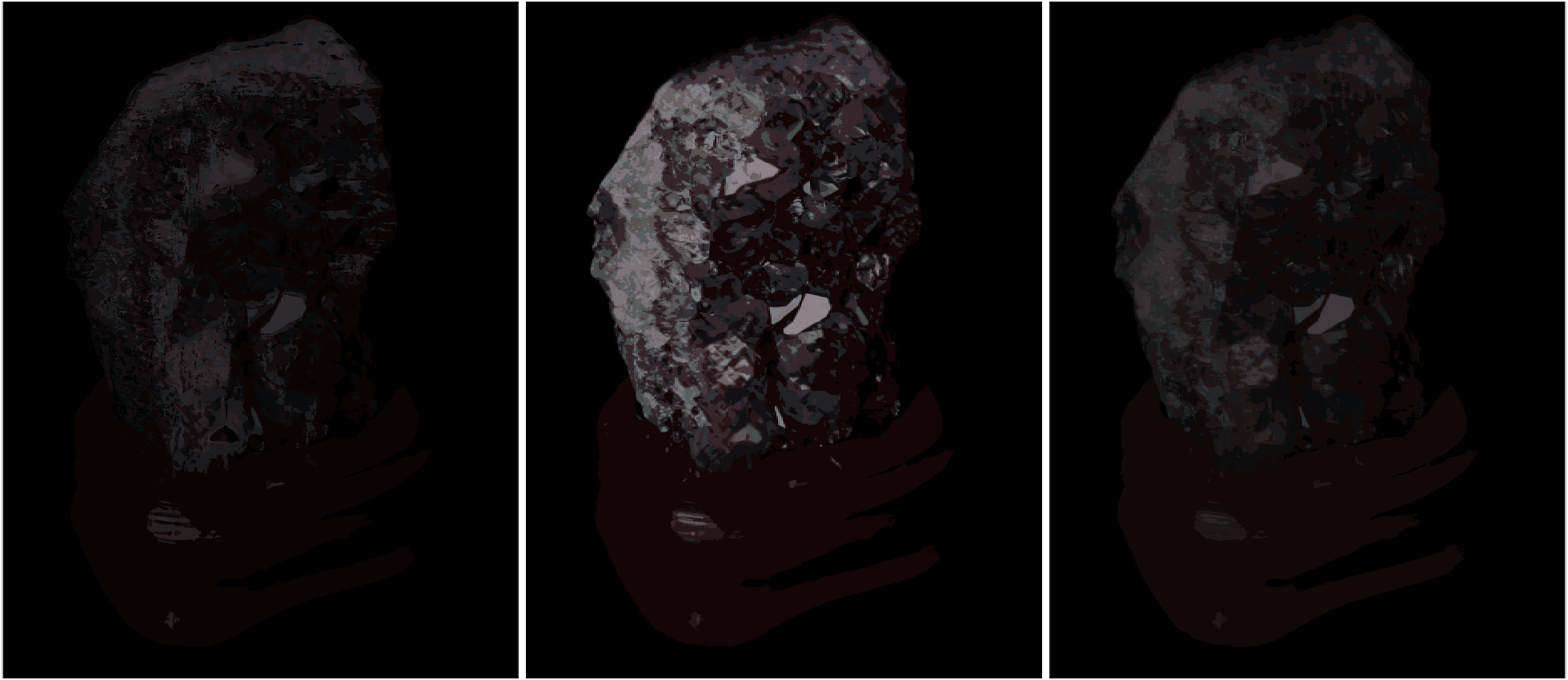}
    \subcaption{$Level = 1$} 
    \end{subfigure} 
    \begin{subfigure}[b]{1\textwidth}
         \centering
         \includegraphics[width=.65\textwidth, height=0.155\textheight]{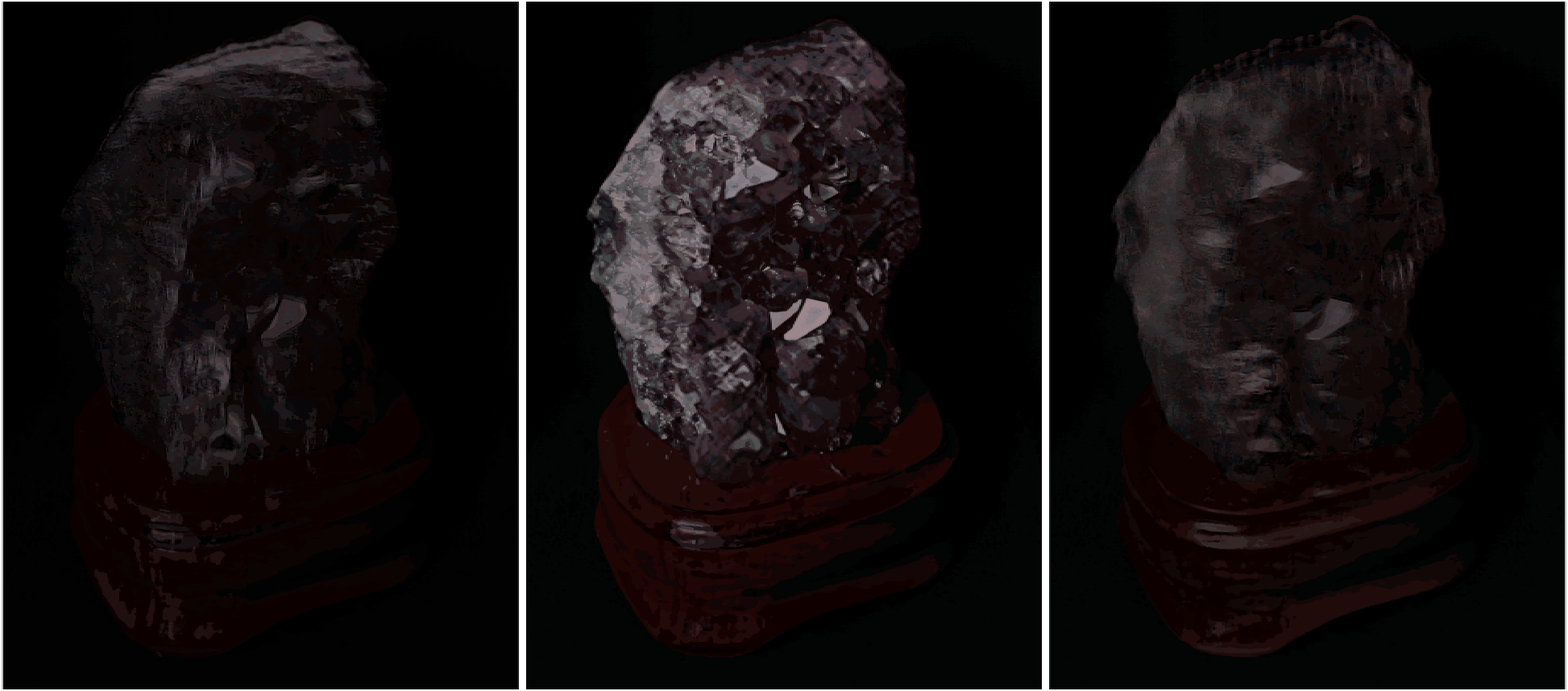}
     \subcaption{$Level = 2$}     
     \end{subfigure}
     \begin{subfigure}[b]{1\textwidth}
         \centering
         \includegraphics[width=.65\textwidth, height=0.155\textheight]{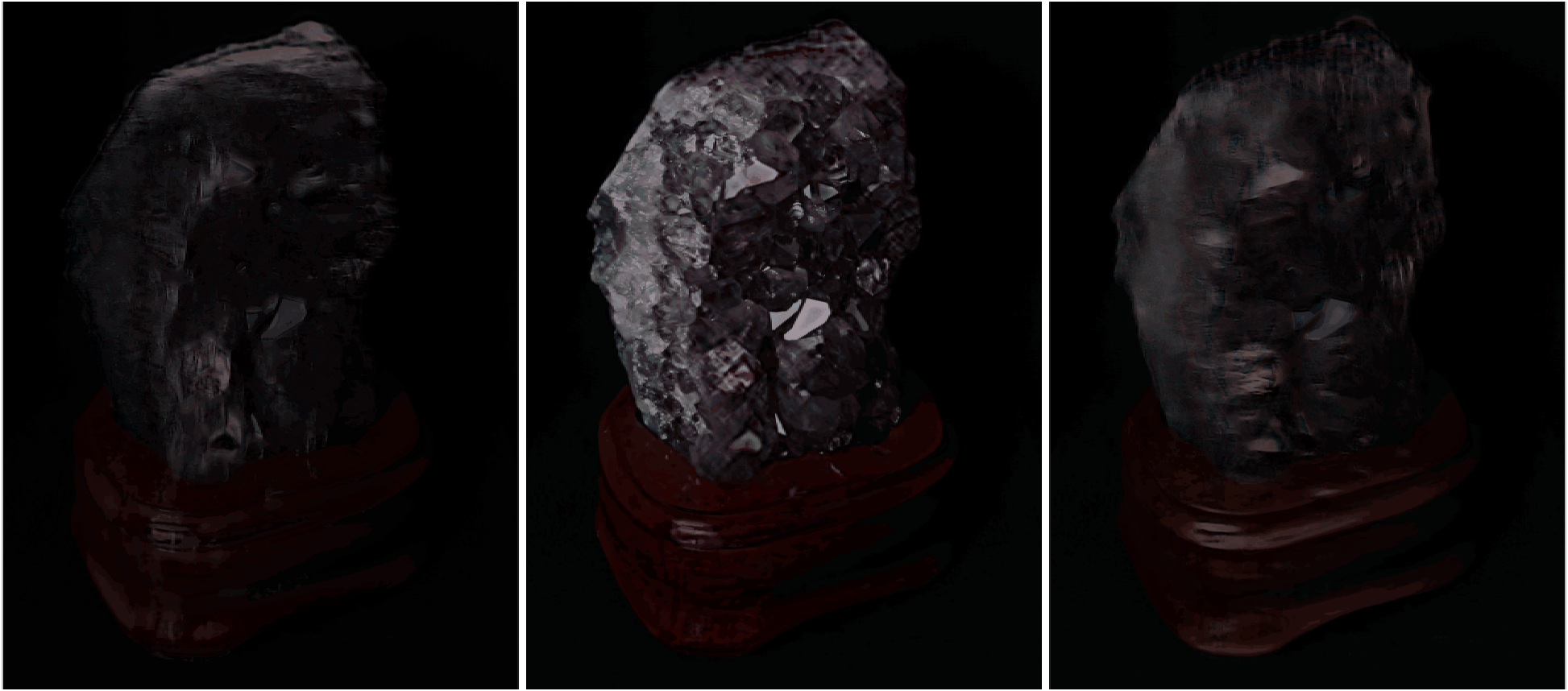}
     \subcaption{$Level = 3$}
     \end{subfigure}  
     \begin{subfigure}[b]{1\textwidth}
         \centering
         \includegraphics[width=.65\textwidth, height=0.155\textheight]{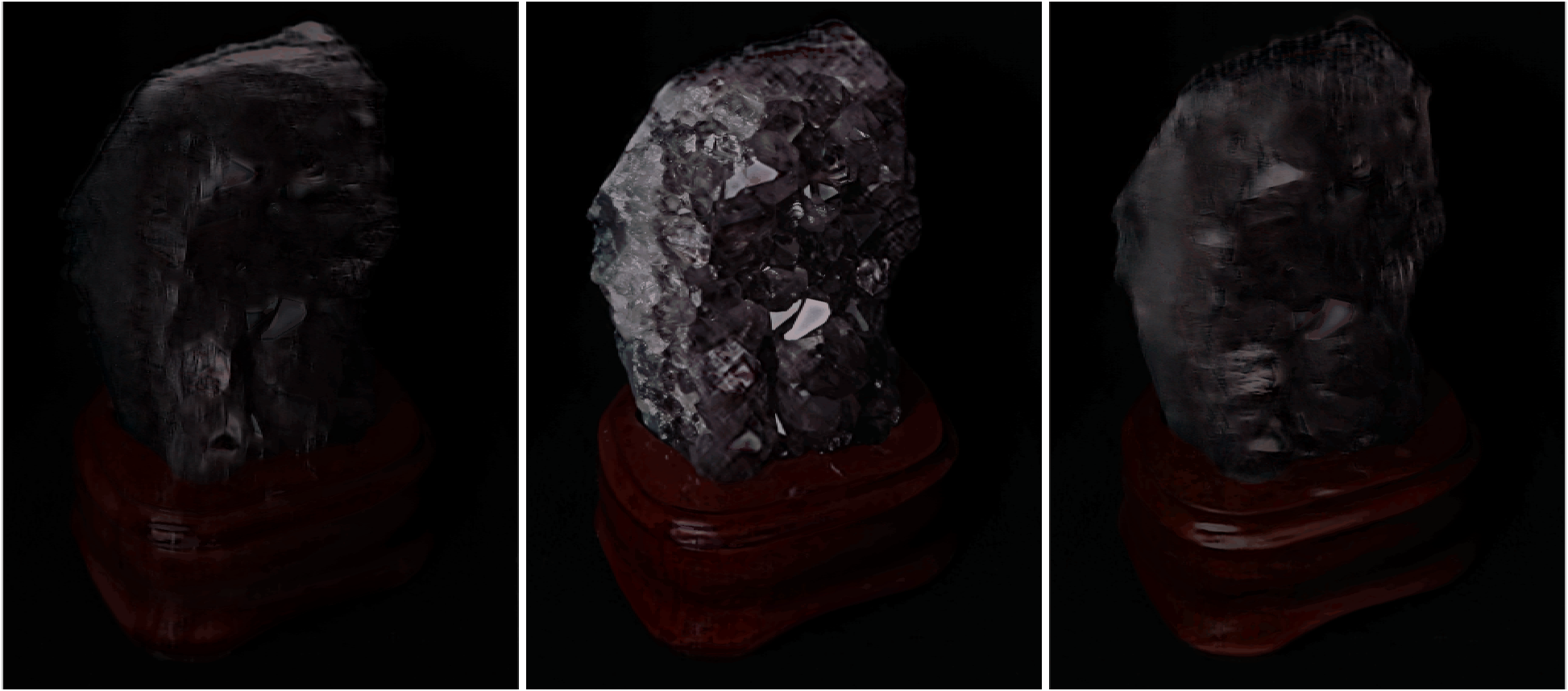}
     \subcaption{$Level = 4$}
     \end{subfigure}  
     \caption{Non-scalable and Scalable layer reconstruction of $Amethyst$.}
\label{fig:reconstruction_amethyst}
\end{figure}

\begin{figure}[!t]
\renewcommand{\figurename}{Figure S}
     \centering
     \begin{subfigure}[b]{1\textwidth}
         \centering
         \includegraphics[width=0.8\textwidth, height=0.155\textheight]{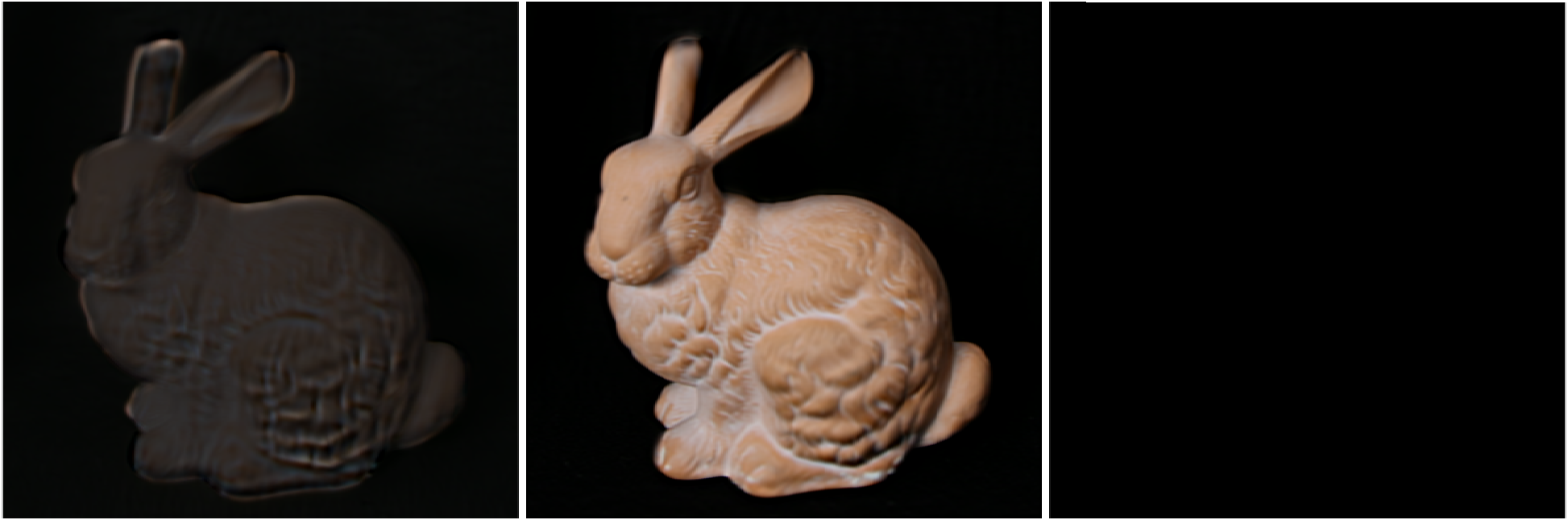}
     \subcaption{Original Layer Pattern}  
     \end{subfigure}
    \begin{subfigure}[b]{1\textwidth}
         \centering
         \includegraphics[width=0.8\textwidth, height=0.155\textheight]{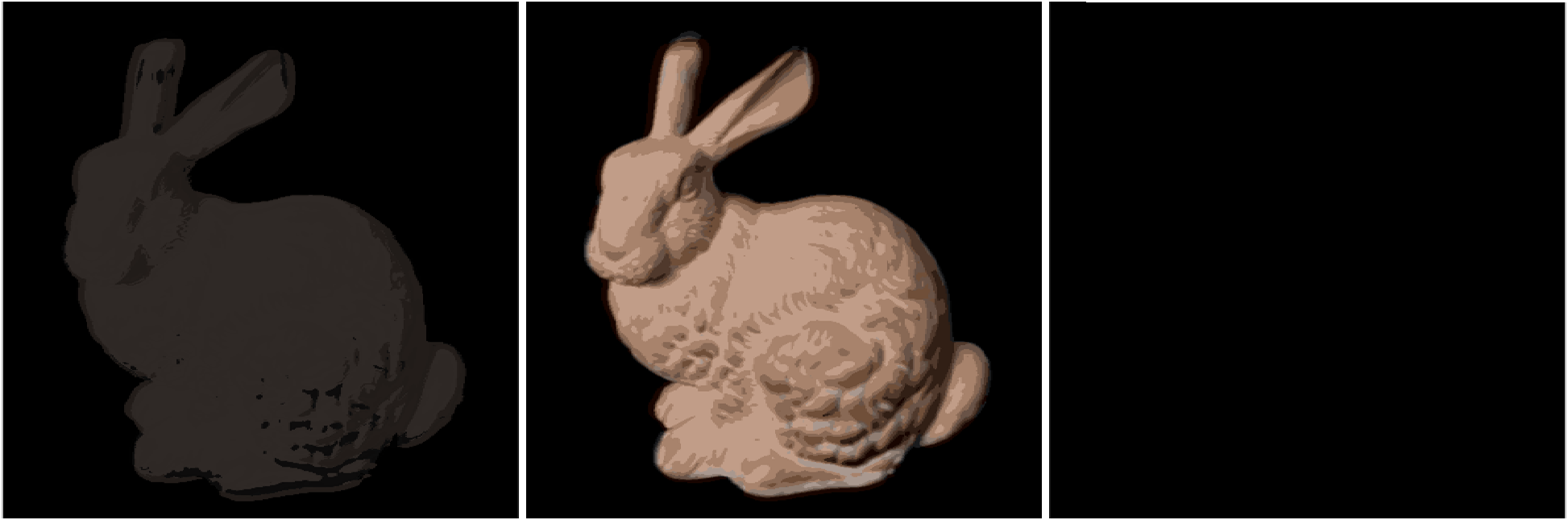}
    \subcaption{$Level = 1$} 
    \end{subfigure} 
    \begin{subfigure}[b]{1\textwidth}
         \centering
         \includegraphics[width=0.8\textwidth, height=0.155\textheight]{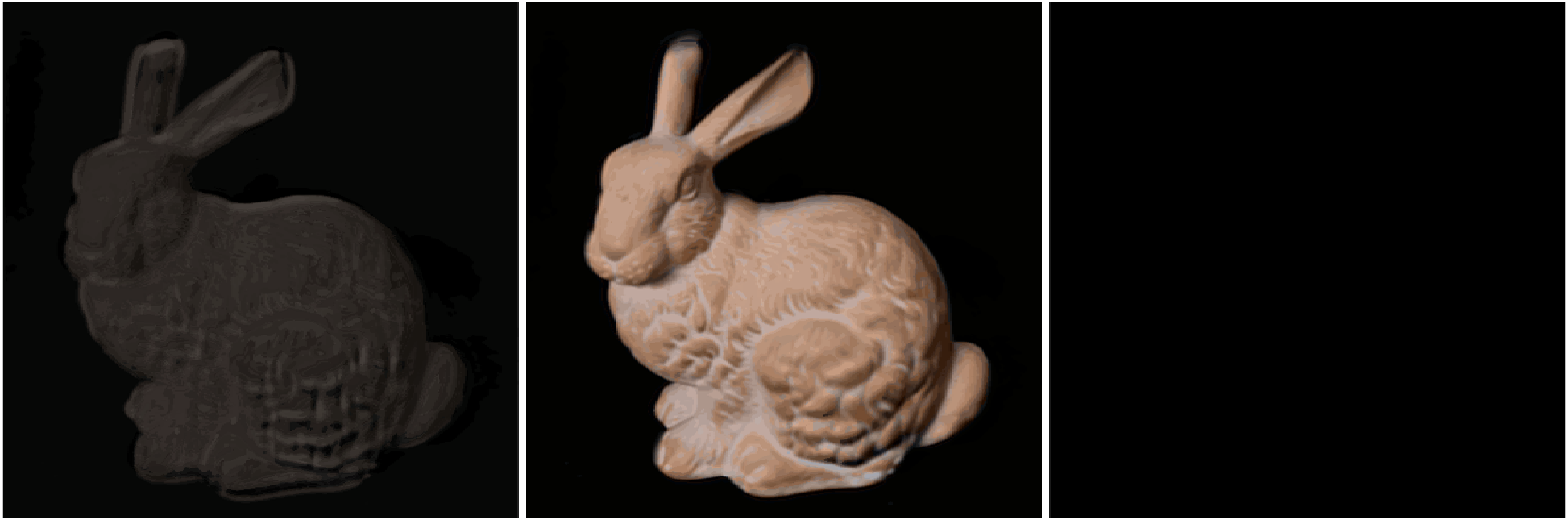}
     \subcaption{$Level = 2$}     
     \end{subfigure}
     \begin{subfigure}[b]{1\textwidth}
         \centering
         \includegraphics[width=0.8\textwidth, height=0.155\textheight]{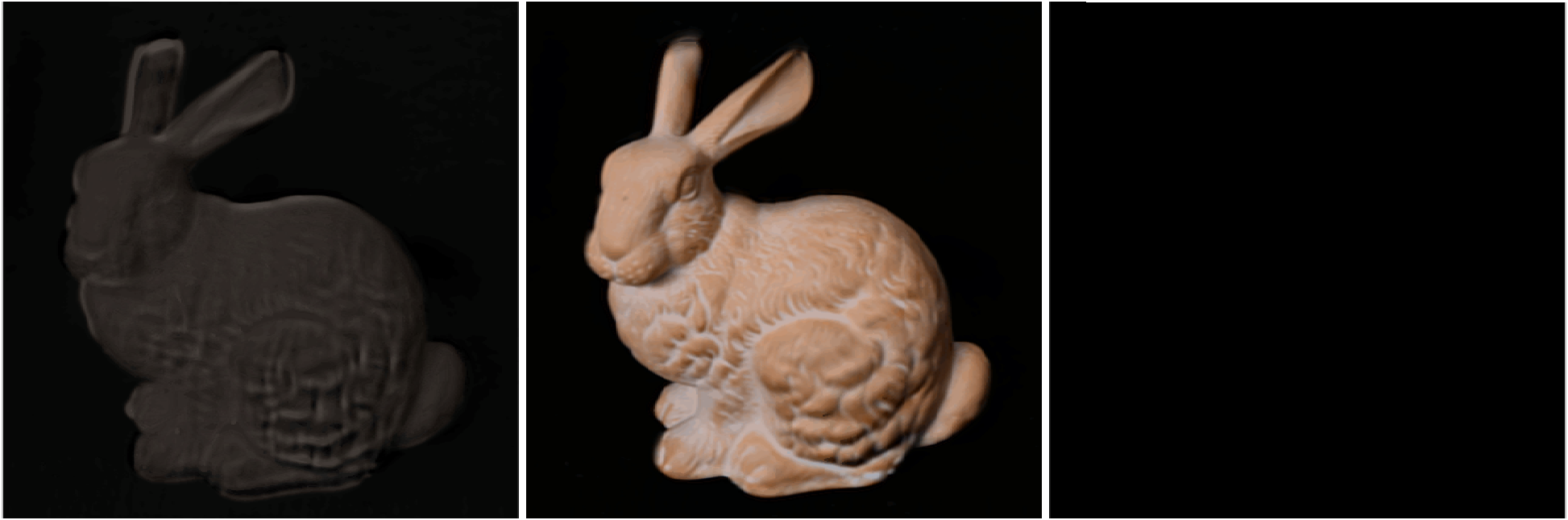}
     \subcaption{$Level = 3$}
     \end{subfigure}  
     \begin{subfigure}[b]{1\textwidth}
         \centering
         \includegraphics[width=0.8\textwidth, height=0.155\textheight]{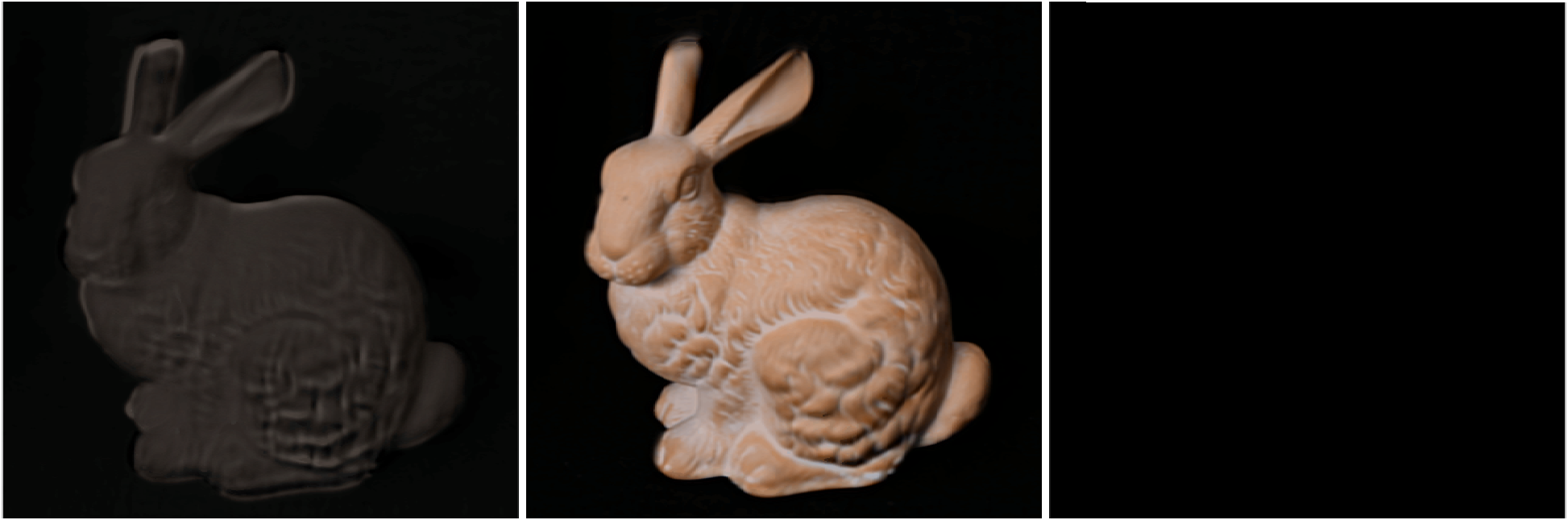}
     \subcaption{$Level = 4$}
     \end{subfigure}  
     \caption{Non-scalable and Scalable layer reconstruction of $Bunny$.}
\label{fig:reconstruction_bunny}
\end{figure}

\begin{figure}[!h]
\renewcommand{\figurename}{Figure S}
     \centering
     \begin{subfigure}[b]{1\textwidth}
         \centering
         \includegraphics[width=0.8\textwidth, height=0.15\textheight]{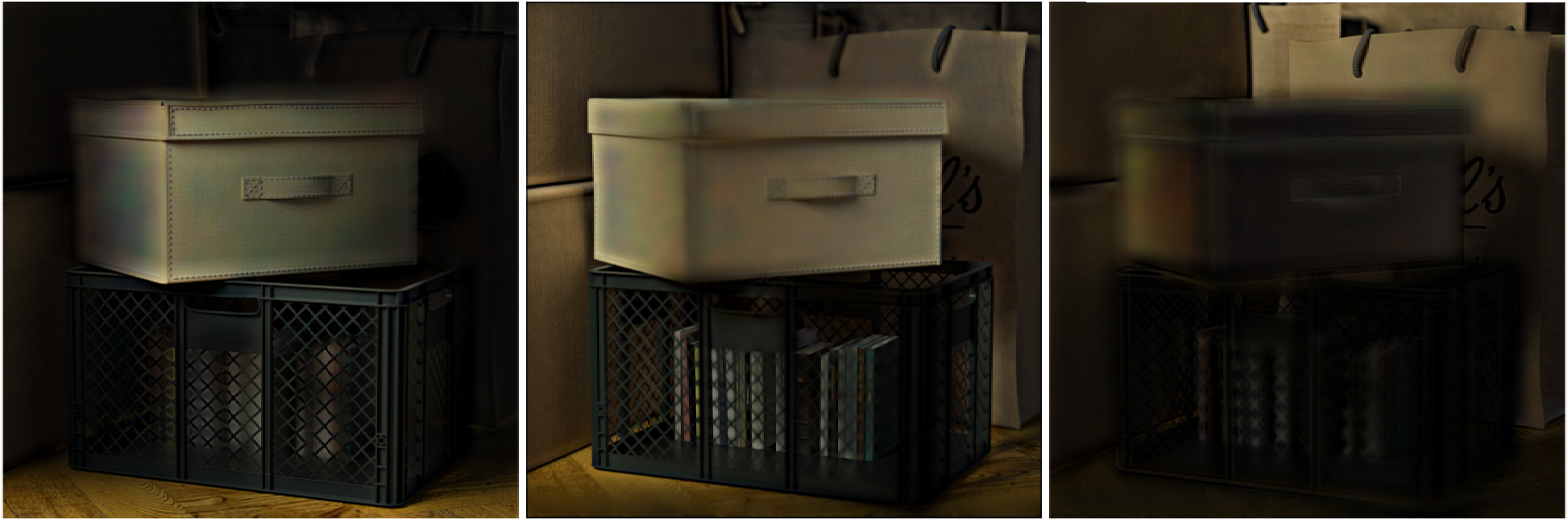}
     \subcaption{Original Layer Pattern}  
     \end{subfigure}
    \begin{subfigure}[b]{1\textwidth}
         \centering
         \includegraphics[width=0.8\textwidth, height=0.15\textheight]{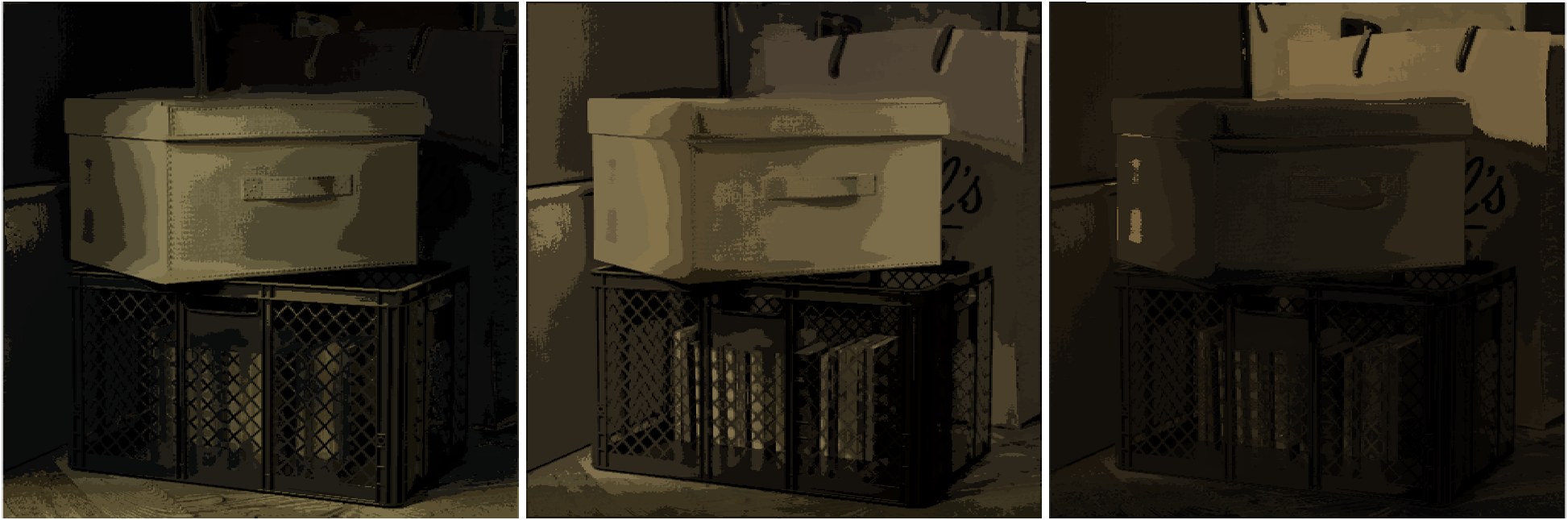}
    \subcaption{$Level = 1$} 
    \end{subfigure} 
    \begin{subfigure}[b]{1\textwidth}
         \centering
         \includegraphics[width=0.8\textwidth, height=0.15\textheight]{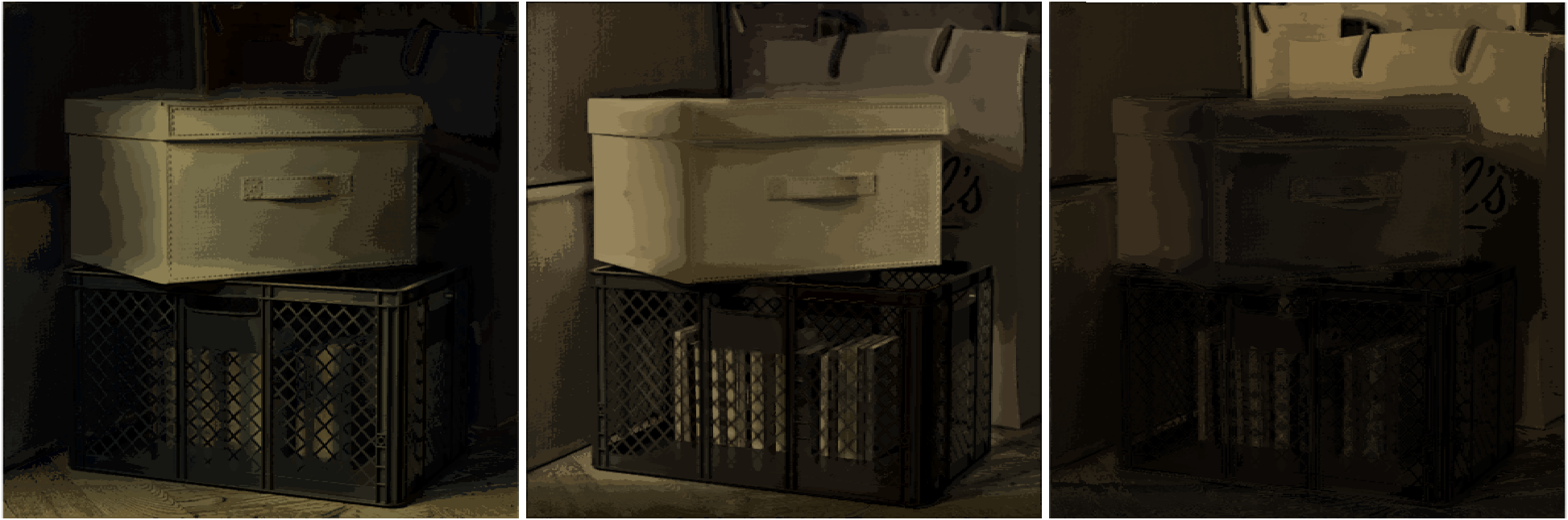}
     \subcaption{$Level = 2$}     
     \end{subfigure}
     \begin{subfigure}[b]{1\textwidth}
         \centering
         \includegraphics[width=0.8\textwidth, height=0.15\textheight]{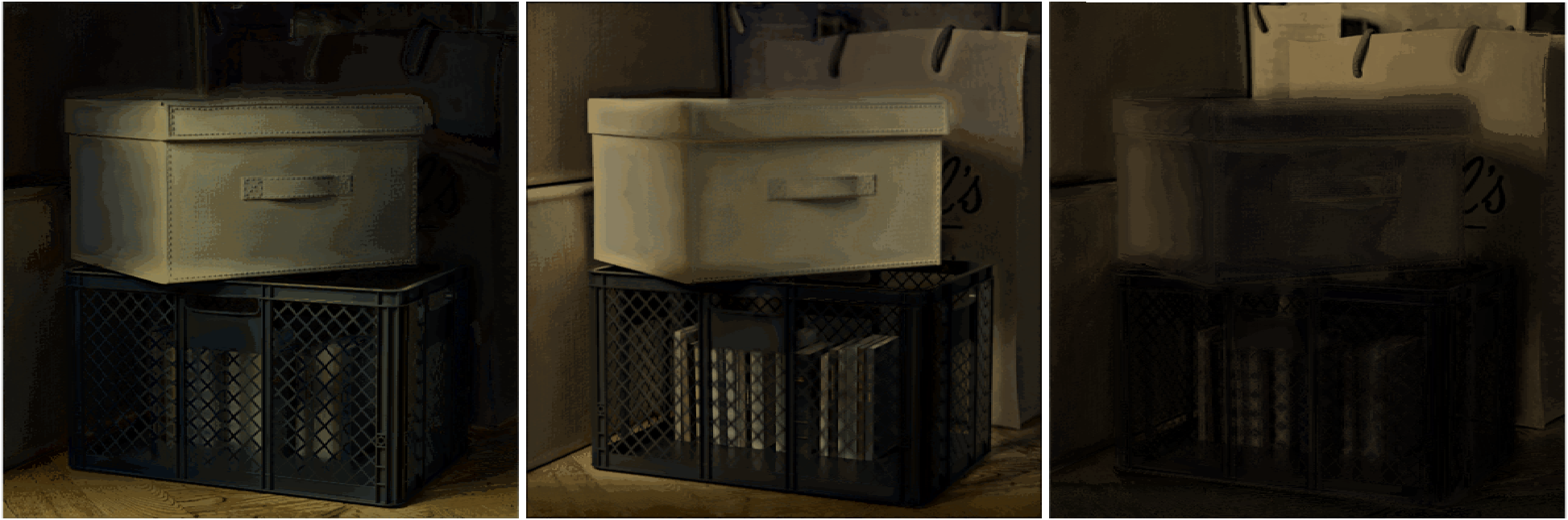}
     \subcaption{$Level = 3$}
     \end{subfigure}  
     \begin{subfigure}[b]{1\textwidth}
         \centering
         \includegraphics[width=0.8\textwidth, height=0.15\textheight]{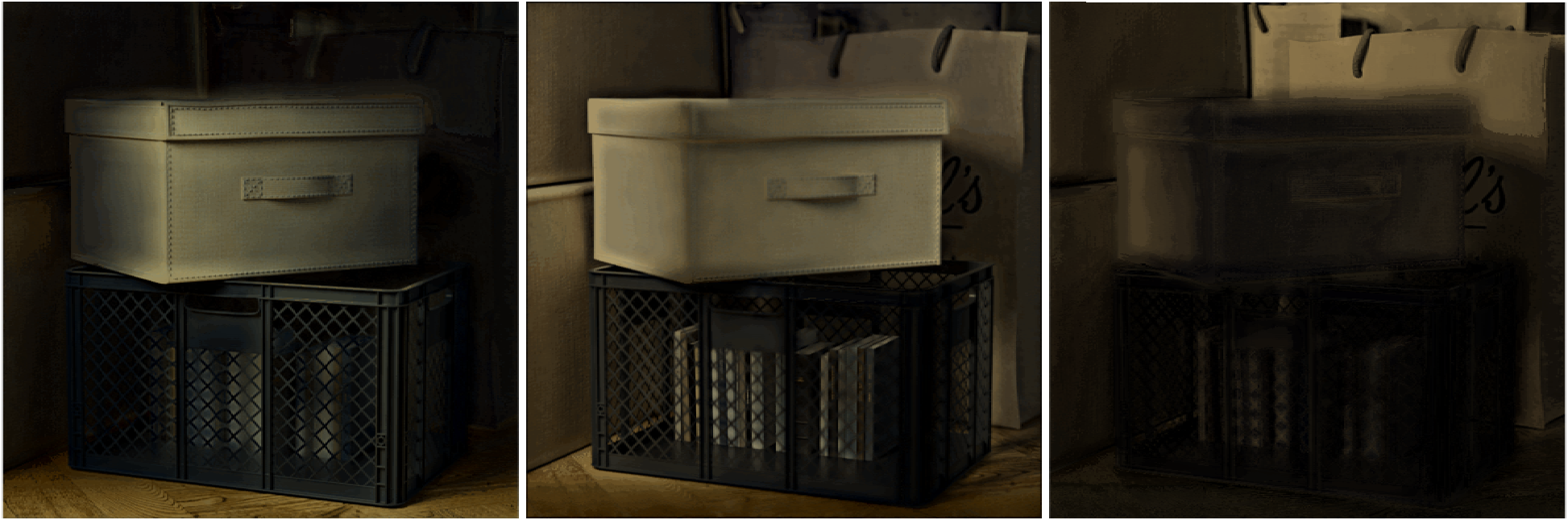}
     \subcaption{$Level = 4$}
     \end{subfigure}  
     \caption{Non-scalable and Scalable layer reconstruction of $Boxes$.}
\label{fig:reconstruction_boxes}
\end{figure}

\begin{figure}[!h]
\renewcommand{\figurename}{Figure S}
     \centering     \begin{subfigure}[b]{1\textwidth}
         \centering
         \includegraphics[width=0.8\textwidth, height=0.15\textheight]{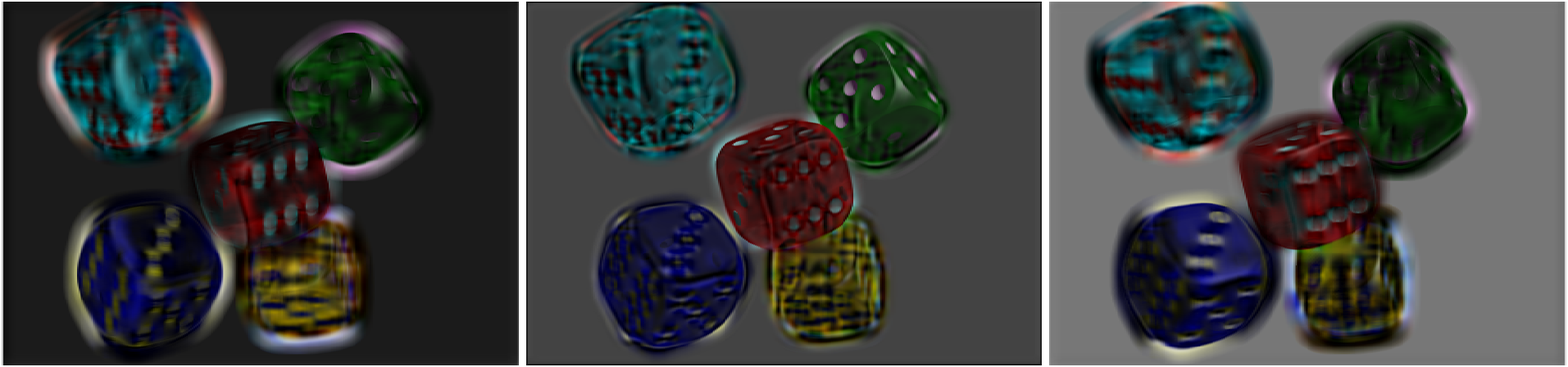}
     \subcaption{Original Layer Pattern}  
     \end{subfigure}
    \begin{subfigure}[b]{1\textwidth}
         \centering
         \includegraphics[width=0.8\textwidth, height=0.15\textheight]{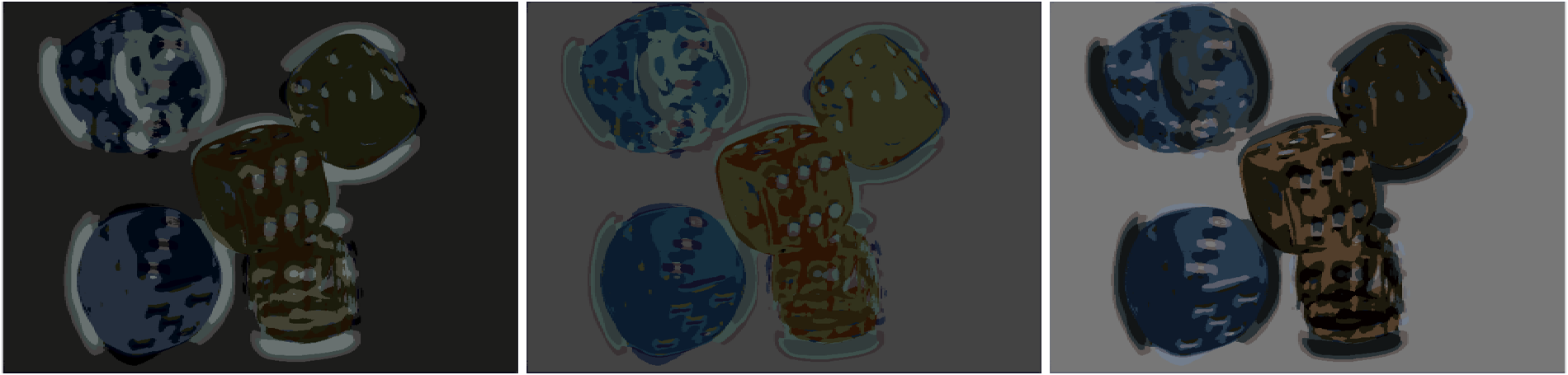}
    \subcaption{$Level = 1$} 
    \end{subfigure} 
    \begin{subfigure}[b]{1\textwidth}
         \centering
         \includegraphics[width=0.8\textwidth, height=0.15\textheight]{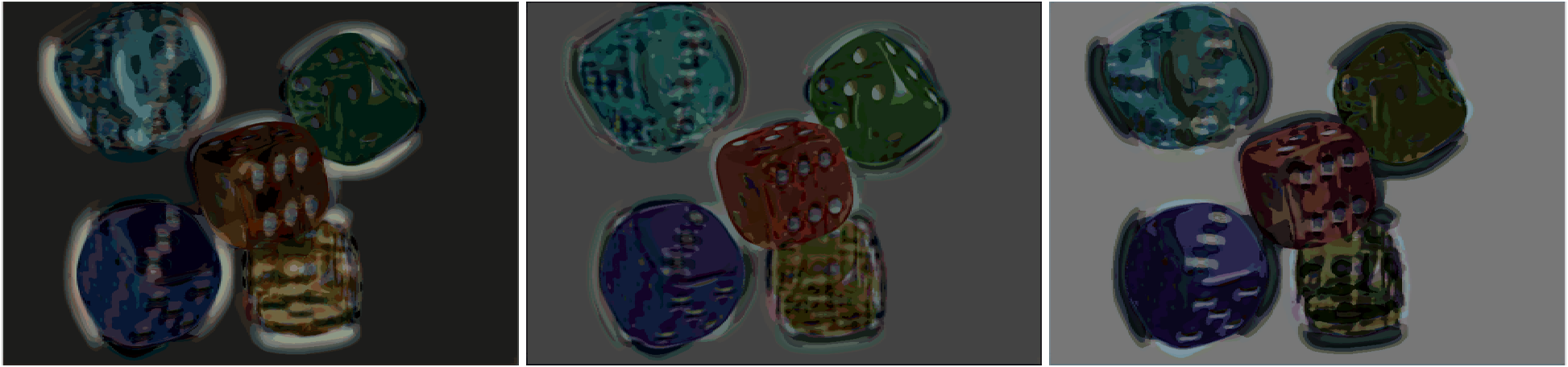}
     \subcaption{$Level = 2$}     
     \end{subfigure}
     \begin{subfigure}[b]{1\textwidth}
         \centering
         \includegraphics[width=0.8\textwidth, height=0.15\textheight]{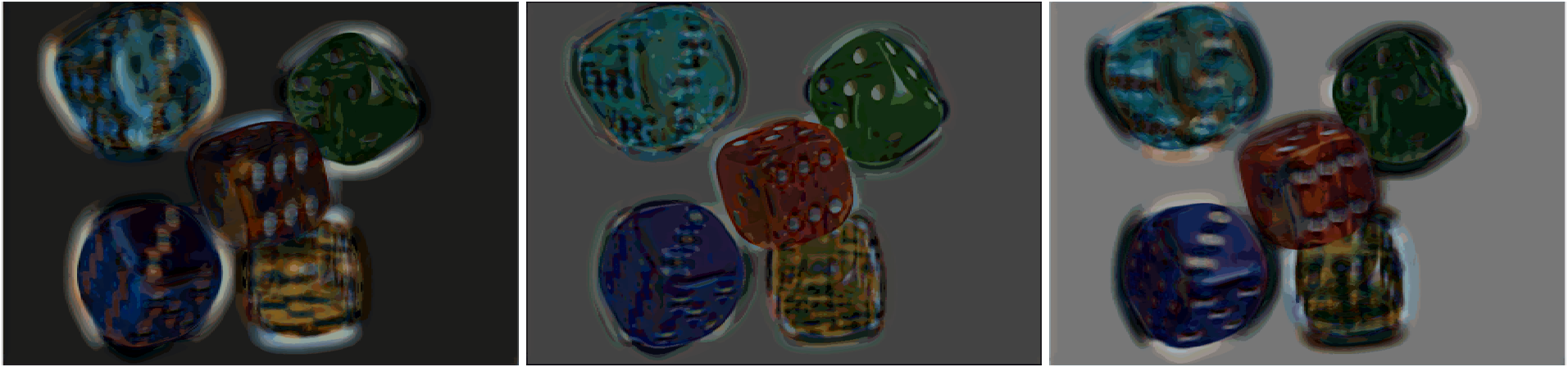}
     \subcaption{$Level = 3$}
     \end{subfigure}  
     \begin{subfigure}[b]{1\textwidth}
         \centering
         \includegraphics[width=0.8\textwidth, height=0.15\textheight]{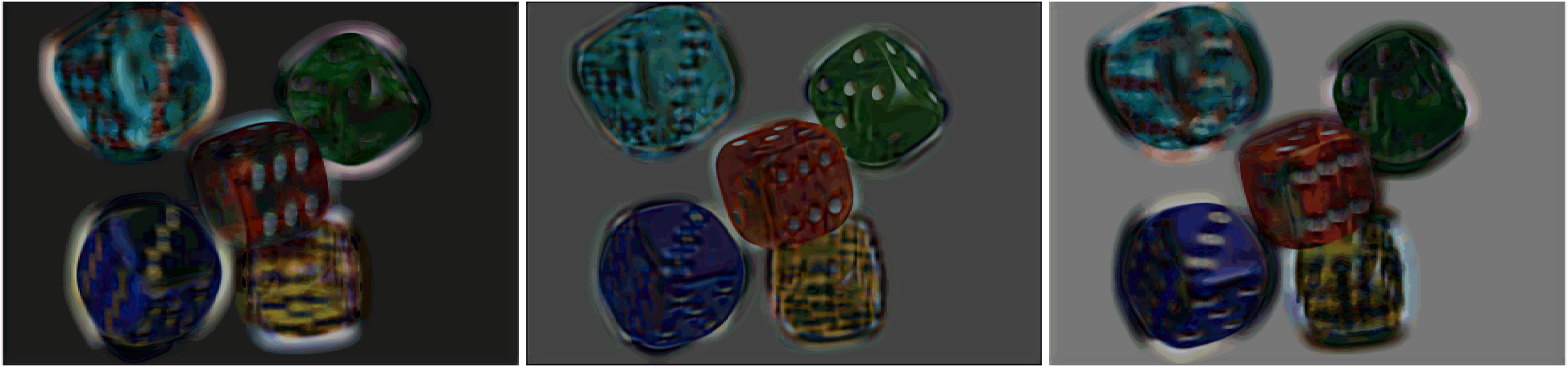}
     \subcaption{$Level = 4$}
     \end{subfigure}  
     \caption{Non-scalable and Scalable layer reconstruction of $Dice$.}
\label{fig:reconstruction_dice}
\end{figure}

\begin{figure}[!h]
\renewcommand{\figurename}{Figure S}
     \centering
     \begin{subfigure}[b]{1\textwidth}
         \centering
         \includegraphics[width=1\textwidth, height=0.13\textheight]{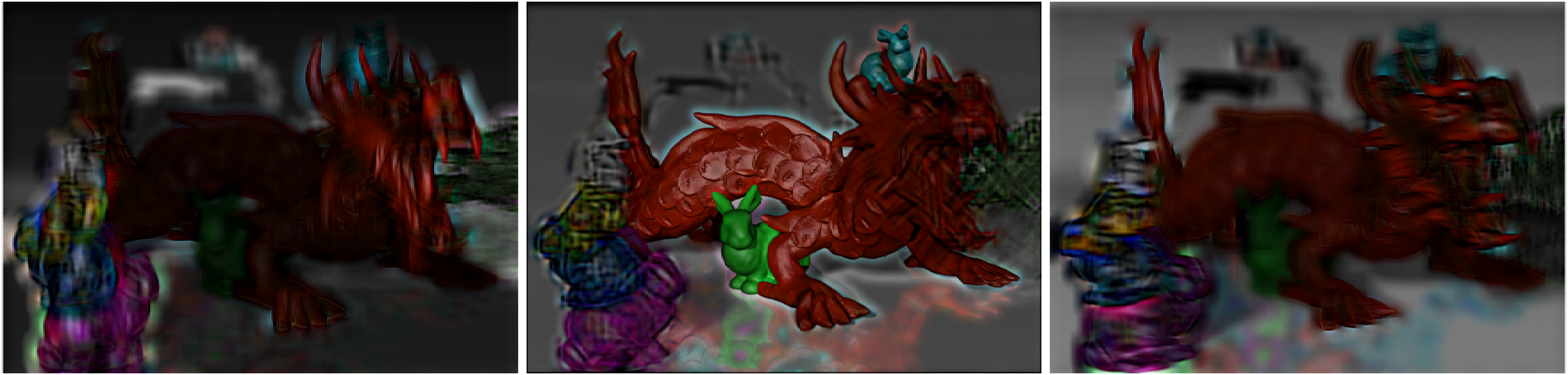}
     \subcaption{Original Layer Pattern}  
     \end{subfigure}
    \begin{subfigure}[b]{1\textwidth}
         \centering
         \includegraphics[width=1\textwidth, height=0.13\textheight]{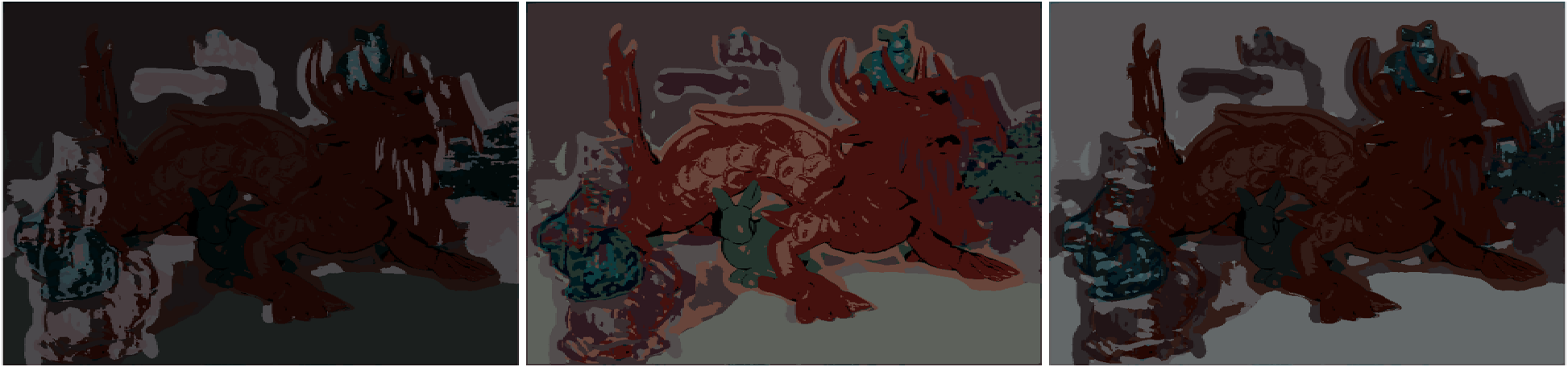}
    \subcaption{$Level = 1$} 
    \end{subfigure} 
    \begin{subfigure}[b]{1\textwidth}
         \centering
         \includegraphics[width=1\textwidth, height=0.13\textheight]{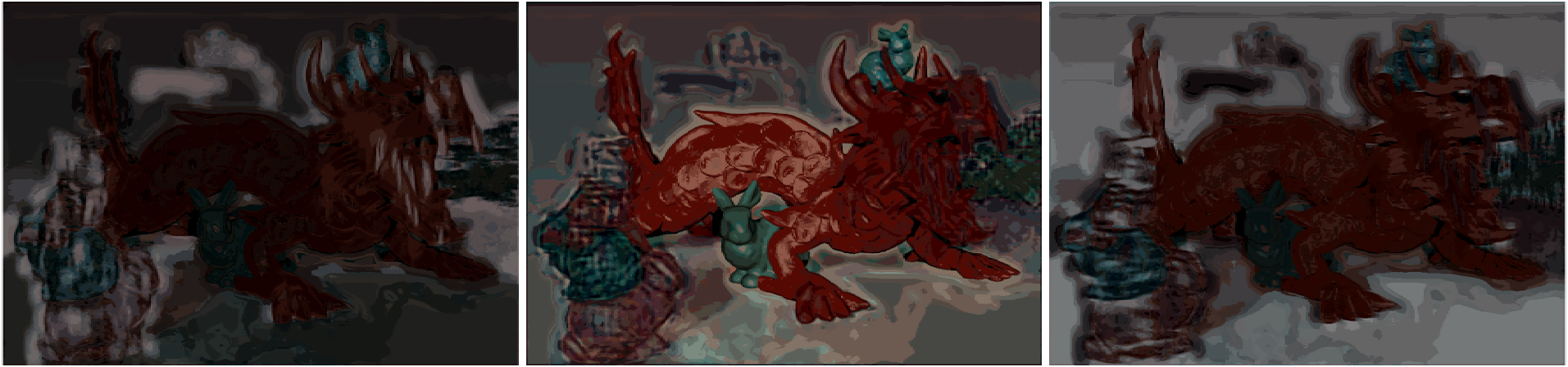}
     \subcaption{$Level = 2$}     
     \end{subfigure}
     \begin{subfigure}[b]{1\textwidth}
         \centering
         \includegraphics[width=1\textwidth, height=0.13\textheight]{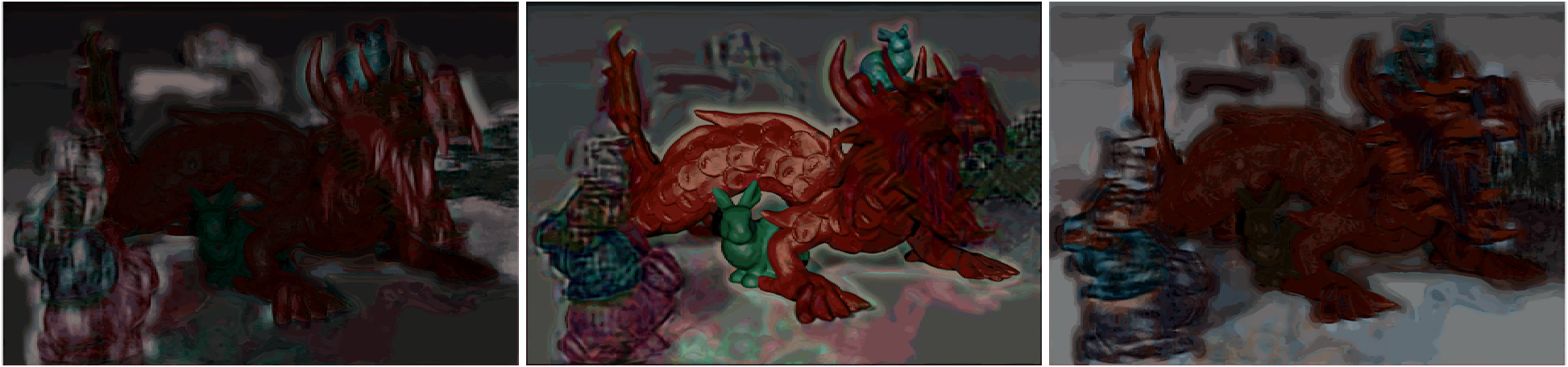}
     \subcaption{$Level = 3$}
     \end{subfigure}  
     \begin{subfigure}[b]{1\textwidth}
         \centering
         \includegraphics[width=1\textwidth, height=0.13\textheight]{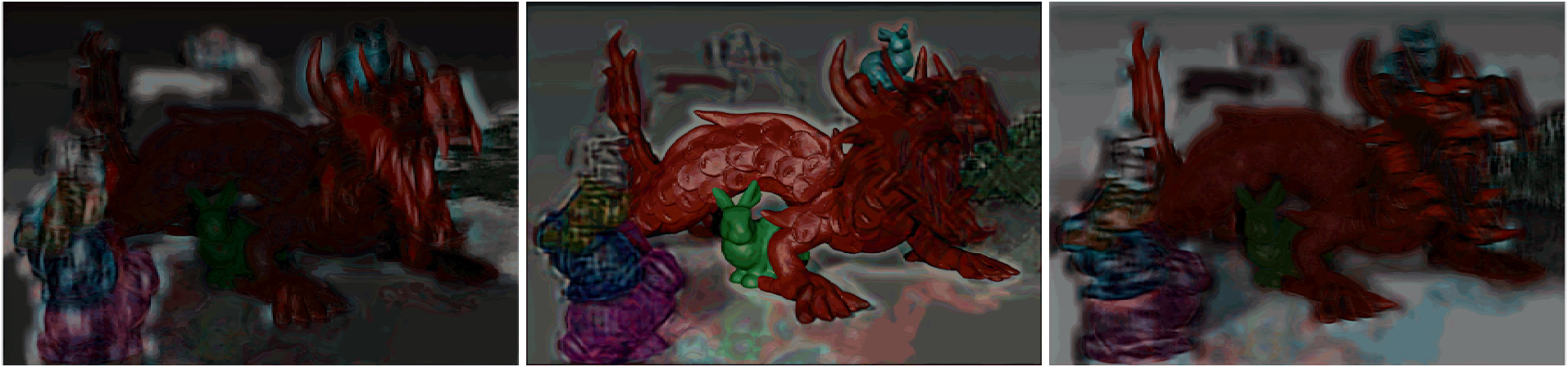}
     \subcaption{$Level = 4$}
     \end{subfigure}  
     \caption{Non-scalable and Scalable layer reconstruction of $DragonAndBunnies$.}
\label{fig:reconstruction_drago}
\end{figure}

\begin{figure}[!t]
\renewcommand{\figurename}{Figure S}
     \centering
     \begin{subfigure}[b]{1\textwidth}
         \centering
         \includegraphics[width=1\textwidth, height=0.13\textheight]{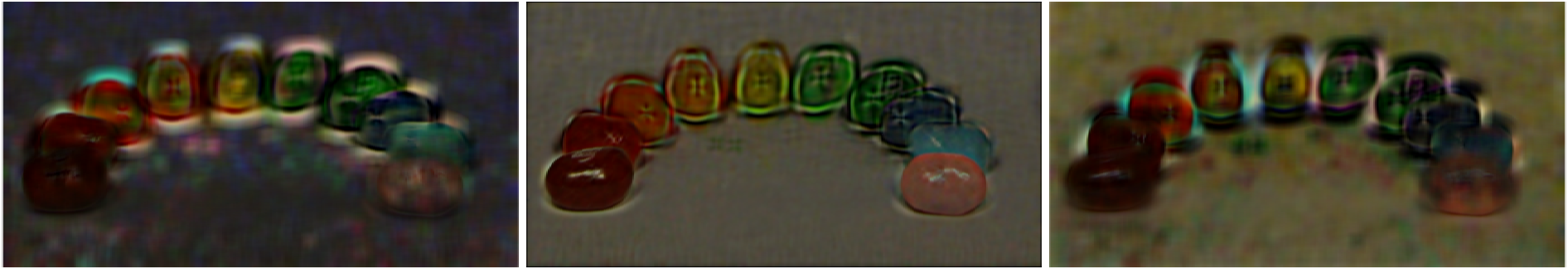}
     \subcaption{Original Layer Pattern}  
     \end{subfigure}
    \begin{subfigure}[b]{1\textwidth}
         \centering
         \includegraphics[width=1\textwidth, height=0.13\textheight]{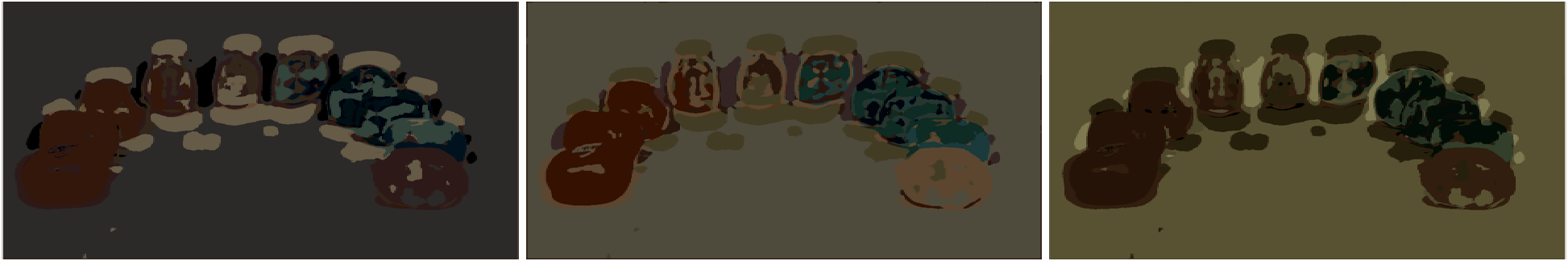}
    \subcaption{$Level = 1$} 
    \end{subfigure} 
    \begin{subfigure}[b]{1\textwidth}
         \centering
         \includegraphics[width=1\textwidth, height=0.13\textheight]{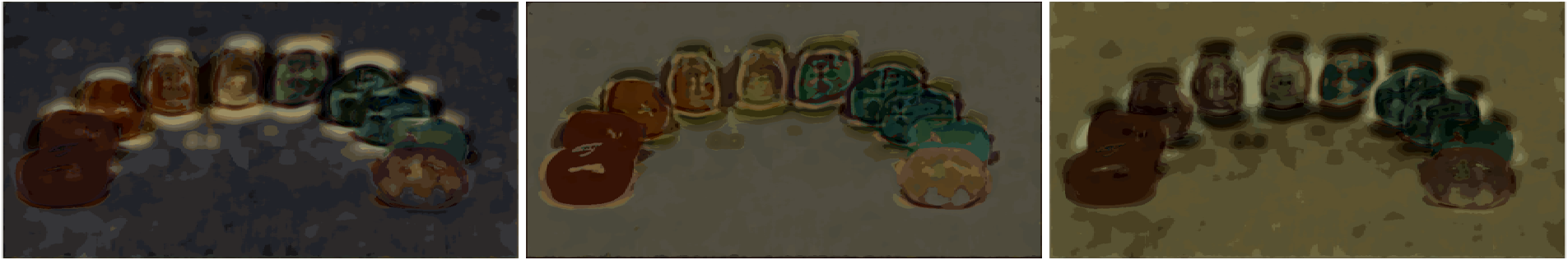}
     \subcaption{$Level = 2$}     
     \end{subfigure}
     \begin{subfigure}[b]{1\textwidth}
         \centering
         \includegraphics[width=1\textwidth, height=0.13\textheight]{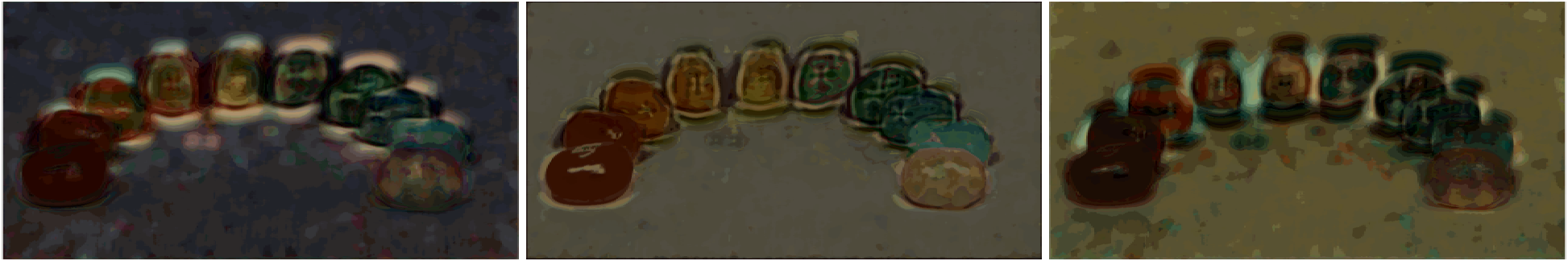}
     \subcaption{$Level = 3$}
     \end{subfigure}  
     \begin{subfigure}[b]{1\textwidth}
         \centering
         \includegraphics[width=1\textwidth, height=0.13\textheight]{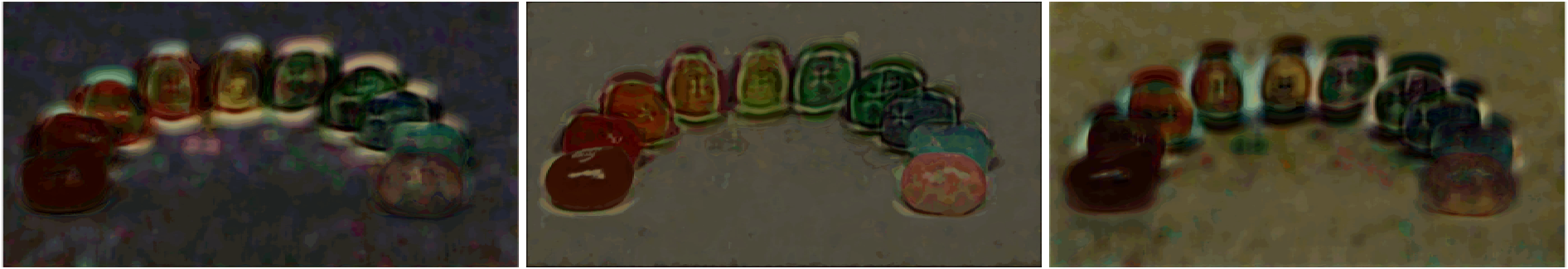}
     \subcaption{$Level = 4$}
     \end{subfigure}  
     \caption{Non-scalable and Scalable layer reconstruction of $Jellybeans$.}
\label{fig:reconstruction_jb}
\end{figure}

\begin{figure*}[!t]
\renewcommand{\figurename}{Figure S}
     \centering
     \begin{subfigure}[b]{0.32\textwidth}
         \centering
         \includegraphics[width=1\textwidth]{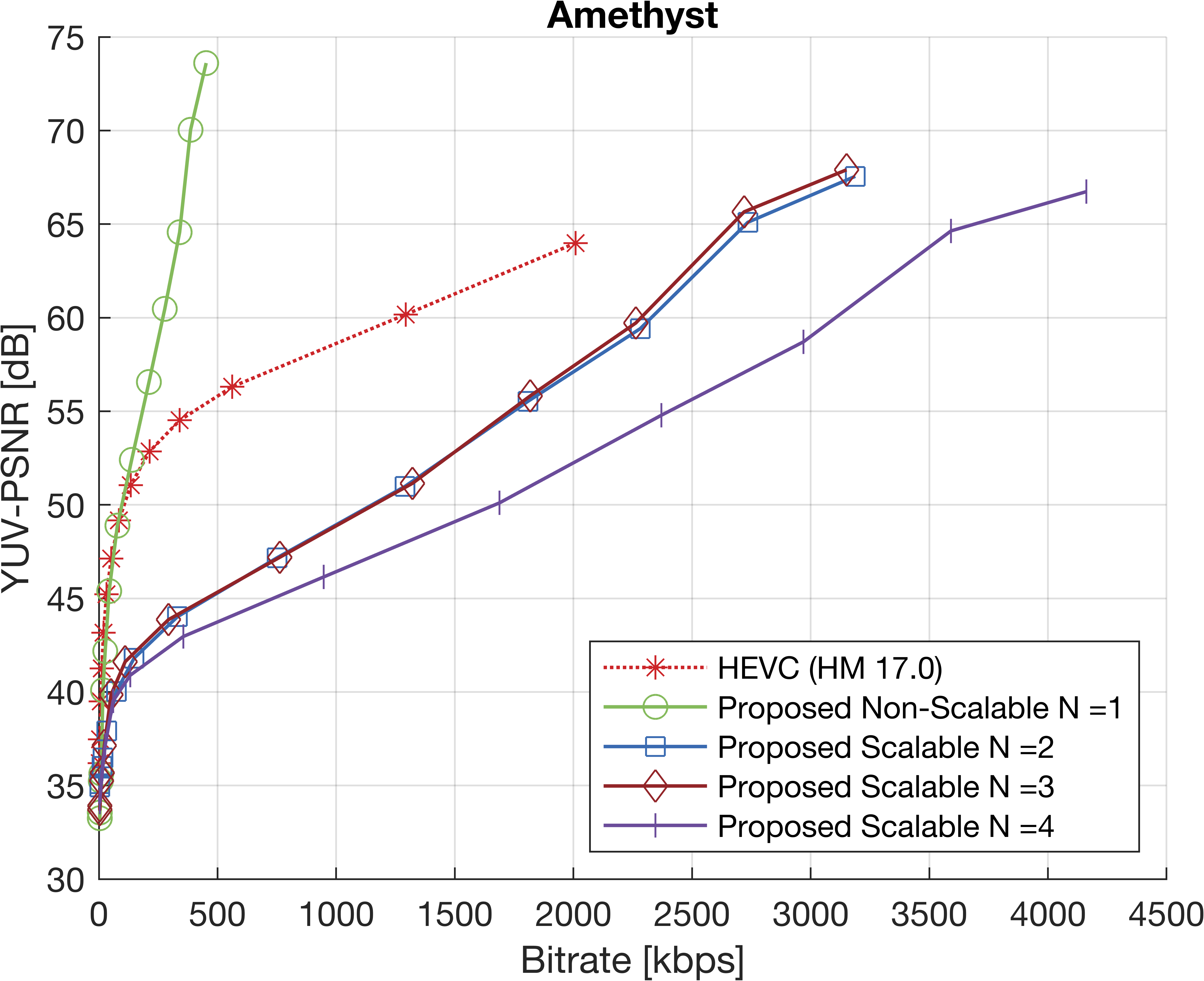}
         \label{fig:amethyst_psnrvsbpp}
     \end{subfigure}
     \begin{subfigure}[b]{0.32\textwidth}
         \centering
         \includegraphics[width=1\textwidth]{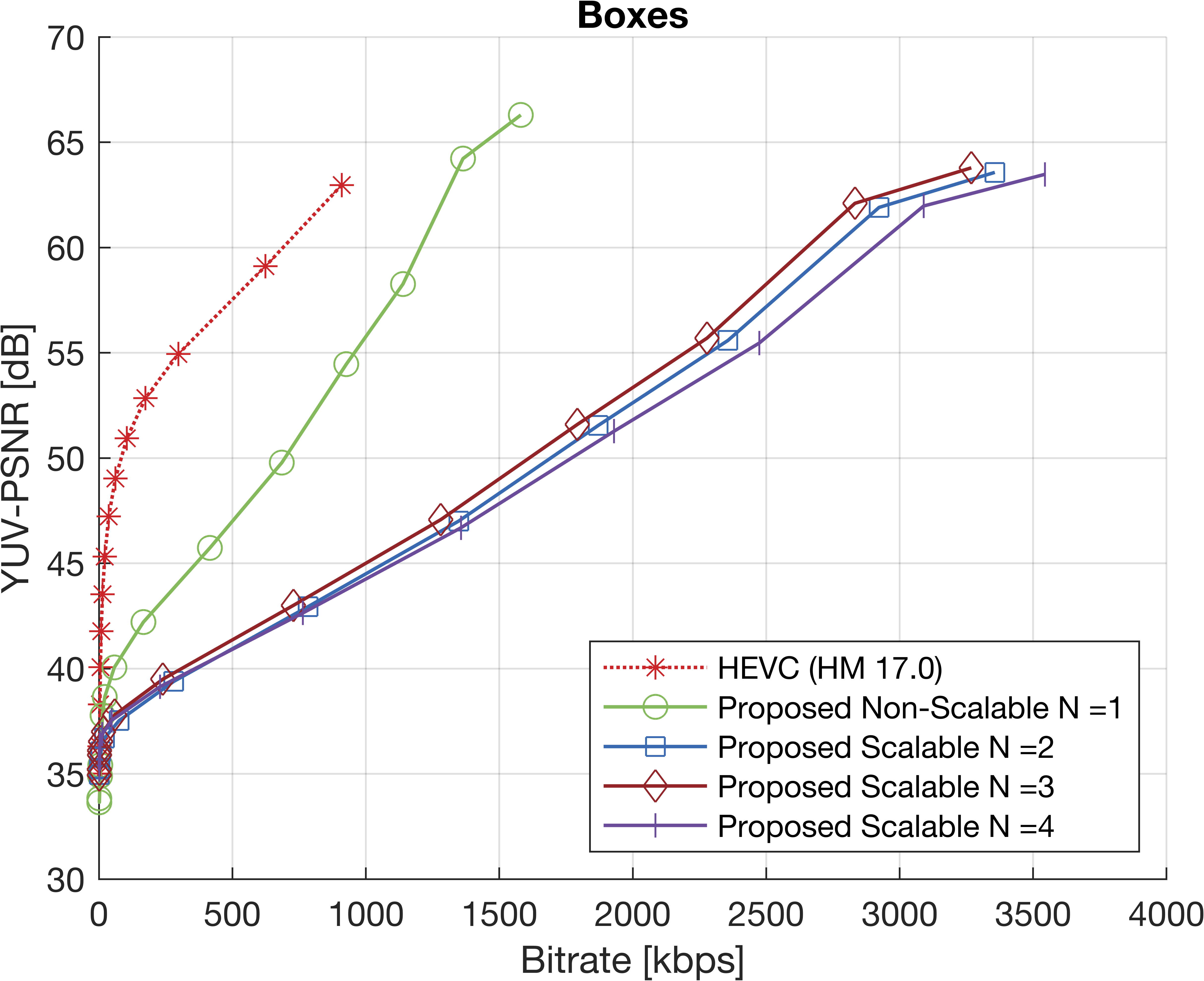}
         \label{fig:boxes_psnrvsbpp}
     \end{subfigure}
     \begin{subfigure}[b]{0.32\textwidth}
         \centering
         \includegraphics[width=1\textwidth]{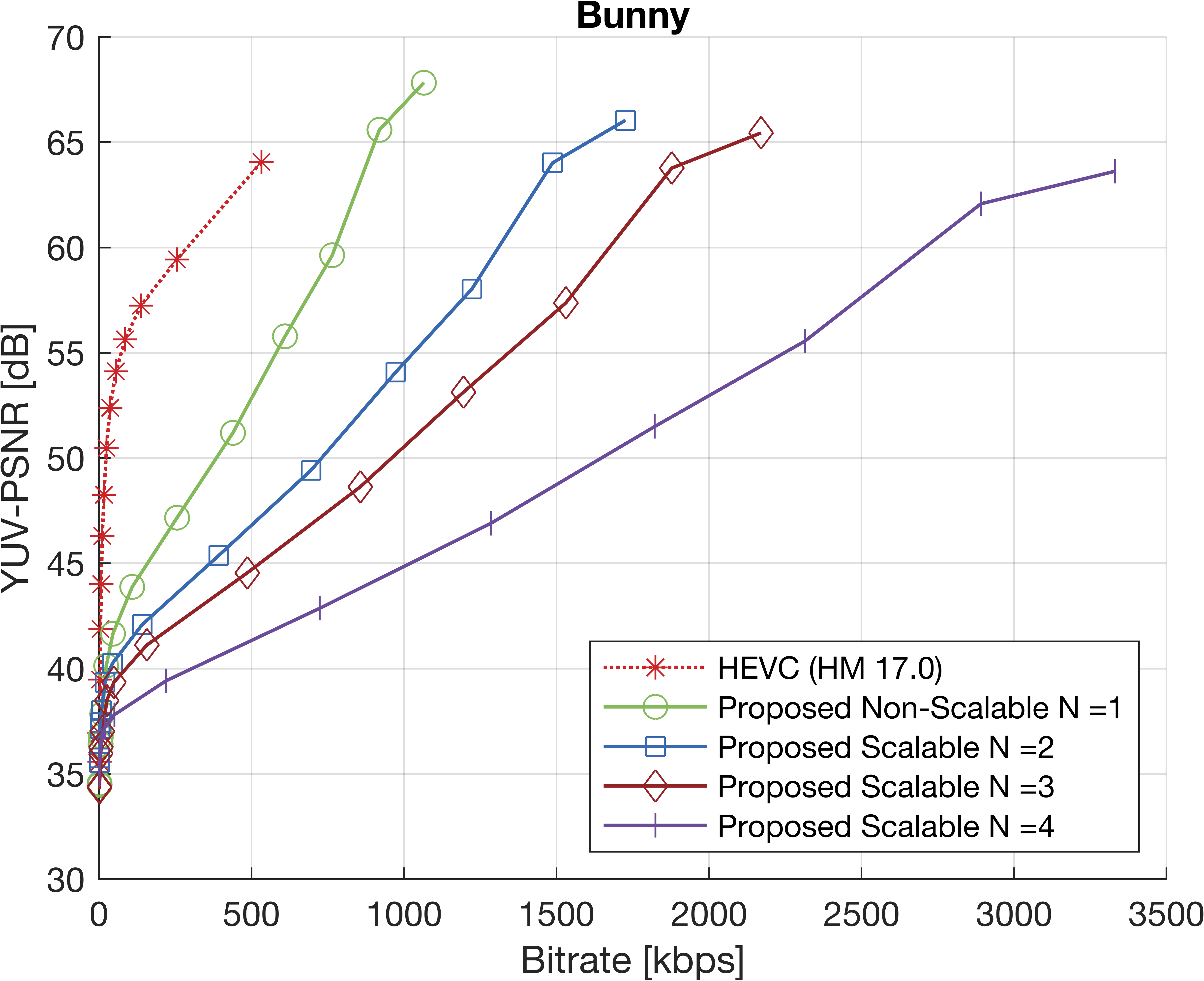}
         \label{fig:bunny_psnrvsbpp}
     \end{subfigure}\\
     \begin{subfigure}[b]{0.32\textwidth}
         \centering
         \includegraphics[width=1\textwidth]{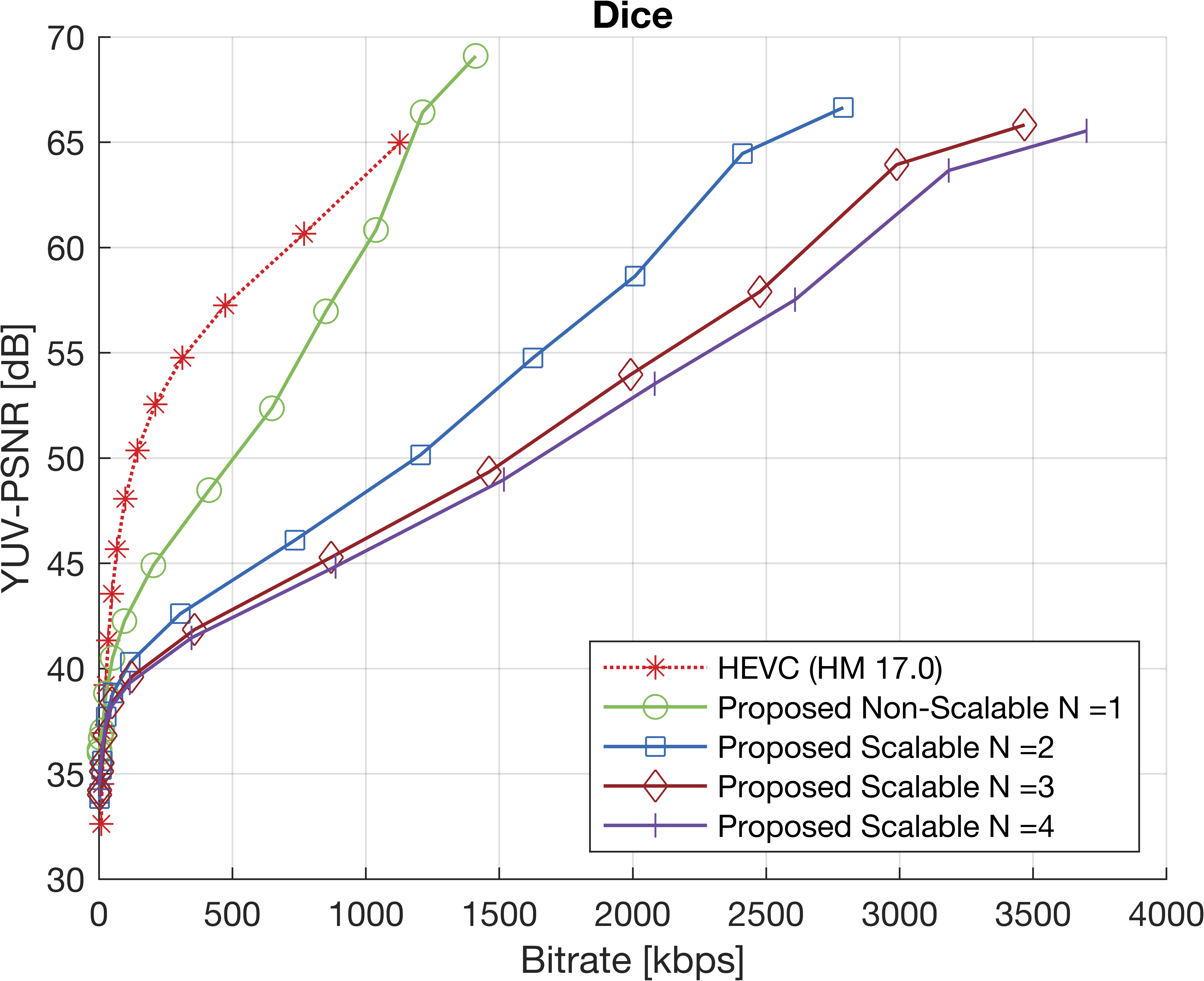}
         \label{fig:dice_psnrvsbpp}
     \end{subfigure}
     \begin{subfigure}[b]{0.32\textwidth}
         \centering
         \includegraphics[width=1\textwidth]{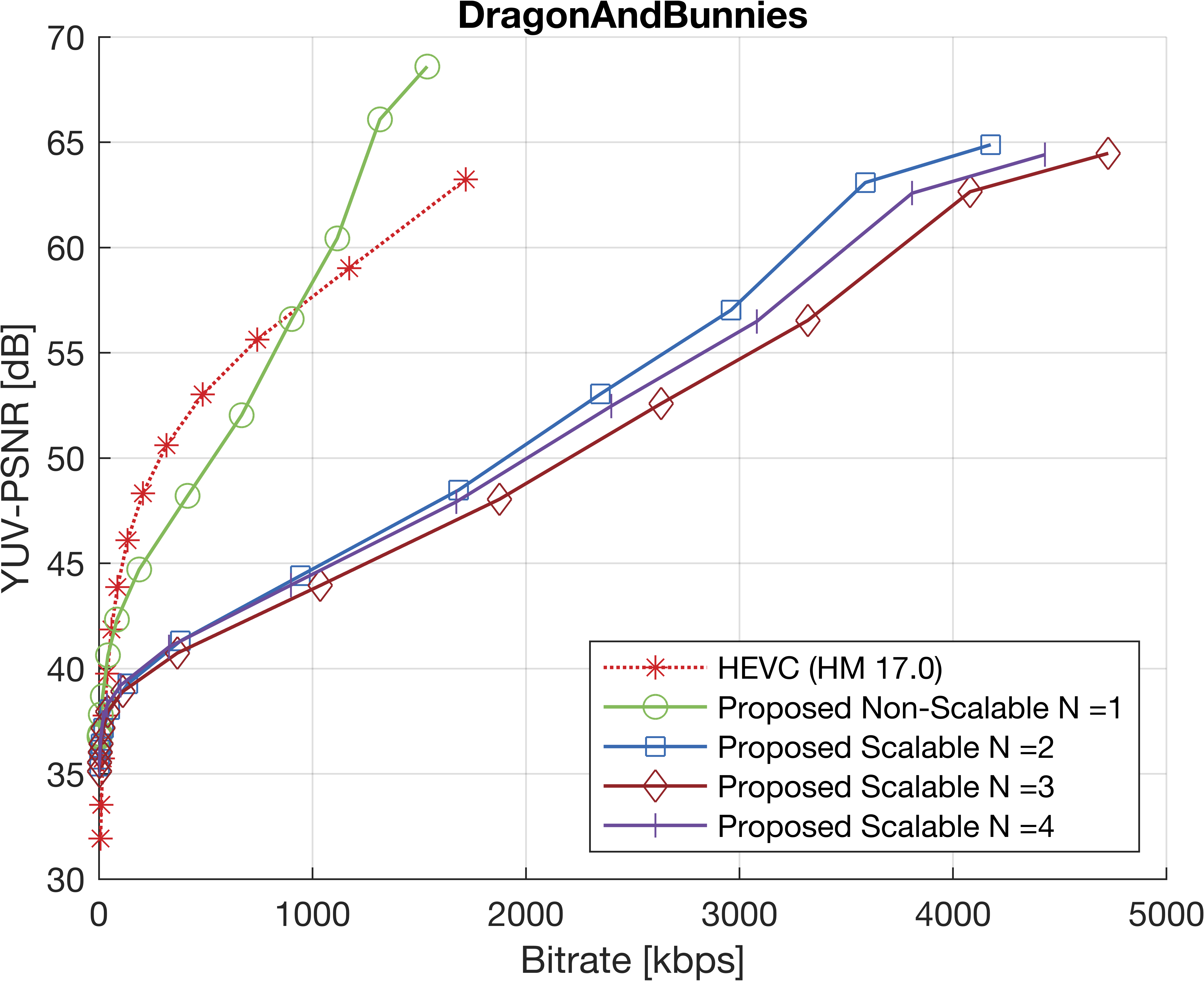}
         \label{fig:drago_psnrvsbpp}
     \end{subfigure}
     \begin{subfigure}[b]{0.32\textwidth}
         \centering
         \includegraphics[width=1\textwidth]{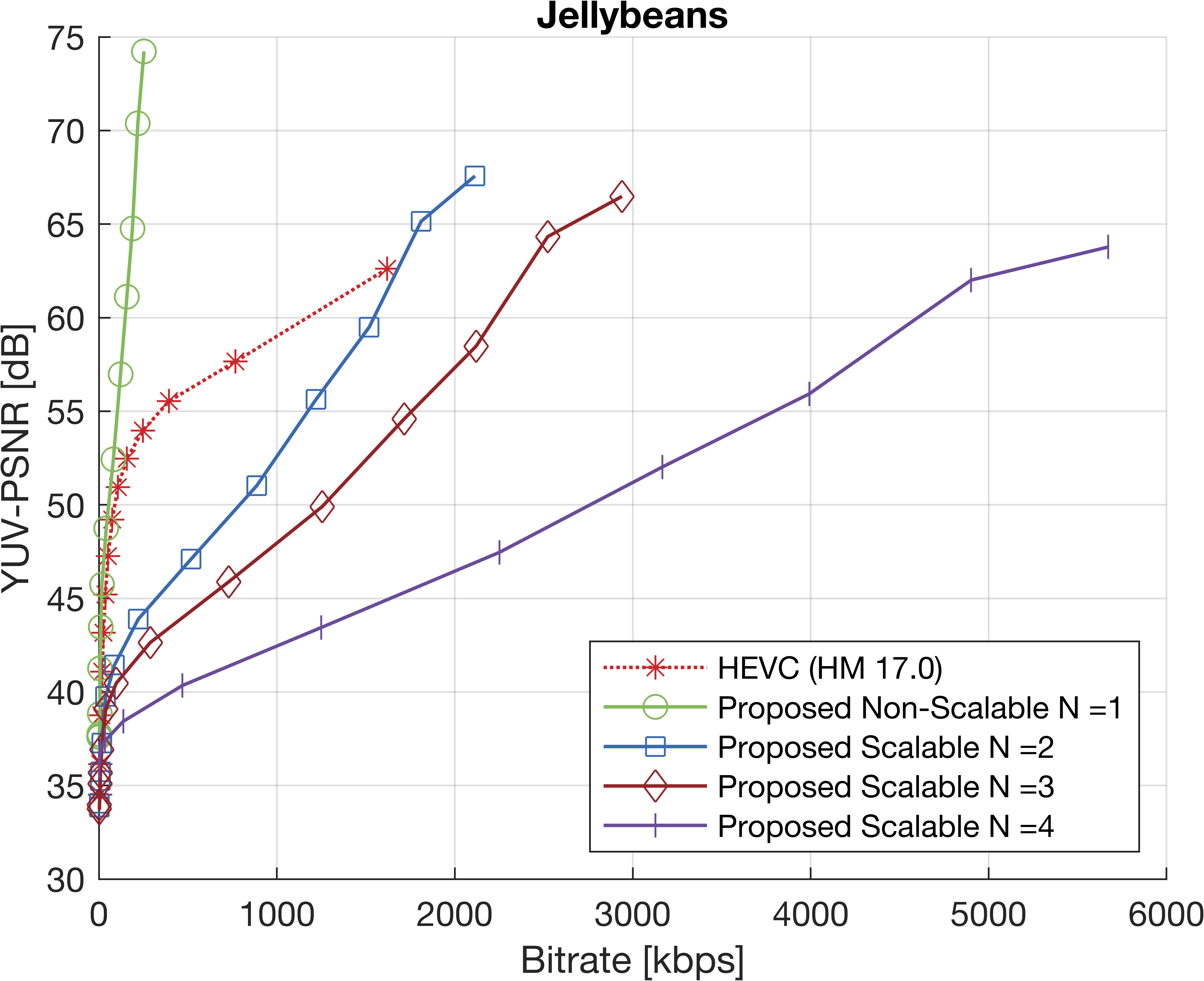}
         \label{fig:Jellybean_psnrvsbpp}
     \end{subfigure}
        \caption{The rate-distortion curve of the proposed compression scheme compared to that of standard video codec (HEVC).}
        \label{fig:RD_curve}
\end{figure*}

The implementation of the proposed scheme is done on a single system with 9th Gen i7, 32 GB RAM, RTX 2080 8 GB Graphics with Ubuntu 22.04 operating system. We implement the sequentially stacked network of 20 2-D convolutional layers in BLOCK I (Fig.~\ref{Pipeline}) using Chainer (version 7.7.0), a Python-based framework. The optimisation problem defined in Equation~\eqref{eq:4} computes the mean square difference between the original and reconstructed light field, which is obtained from additive layers optimised by the CNN. These layer patterns are represented as weighted binary images, which can optimally approximate them. We perform experiments considering various scalable layers and number of binary images as depicted Fig.~\ref{scalablepipeline}. As we incorporate more layers into the encoding process, the reconstruction quality increases progressively. This validates the progressive increment of reconstruction quality as more information is utilised as shown in Fig.~S\ref{fig:scalable_non_scalable_psnr_size}. Here, we introduce rate scalability aspect in the framework. The DBN encodes weighted binary patterns into a dimensionally reduced representation. The encoder (Fig.~\ref{DBN}) has layer size of $F_1-F_2-F_3-F_4$ and symmetric decoder layout $F_4-F_5-F_6-F_7$, where $F_1=F_7,\ F_2=F_6\ \text{and}\ F_3=F_5$. The training sample for the network is a set of $64\times64$ 2-D image blocks derived from the exact locations in all image views of the sample light field. The sample patch of $64\times64$ pixels is collected from the training data $32$ pixels apart in horizontal and vertical directions, discarding the patches with nearly uniform intensities. The training samples are collected from several light field datasets~\cite{vaish2008new, marwah2013compressive, honauer2017dataset}. The training data differs from the test data. The testing is done on \textit{Amethyst}, \textit{Boxes}, \textit{Bunny}, \textit{Dice}, \textit{DragonsAndBunnies} and \textit{Jellybeans} light field dataset having $5\times5$ viewpoints. While training, eight sets were considered, each consisting of 64000 samples and trained for 30 epochs taking around 5 hours for each sample set. The initial weights are small random values procured from a normal distribution having zero mean and a standard deviation of 0.1. We adjust the weights with a learning rate of 0.1 after each mini-batch (Equation~\ref{eq:energy_rbm}). To encourage stability and prevent overfitting, we use a regularization technique involving a combination of additive and penalty terms. Specifically, we add 0.9 times the value of the previous update to each weight, which helped to ensure gradual changes over time. Additionally, we subtract 0.00002 times the weight value from the update, which penalises large weights and encouraged the model to focus on essential features. The latent data from BLOCK II is encoded using HEVC (HM 17.0), generating an encoded bit-stream that can be scaled to reconstruct at any desired bitrate (quality). The performance of our proposed Scalable and Non-Scalable coding schemes are compared with standard video codec, HEVC (HM 17.0)~\cite{sullivan2012overview} as illustrated in Fig.~\ref{fig:RD_curve}. We covered the range of HEVC (main10) quantization parameters, \textit{QP} $0$, $4$, $8$, $12$, $16$, $20$, $24$, $28$, $32$, $36$, $40$, $44$, $48$, $51$ to test the performance at both high and low bitrate cases. In Table~\ref{tab:bdratepsnr}, an objective assessment to compare bitrate reduction of the proposed scheme (Scalable and Non-Scalable) with respect to HEVC codec is done using the Bjontegaard~\cite{bjontegaard2001calculation} metric. The more negative value of BD-Rate, the more gain in bitrate is achieved by the proposed scheme and similarly for BD-PSNR. Subsequently, the decoder of DBN decompresses the bitstream from BLOCK II to regenerate the approximated layer patterns and then transform them into the whole light field. In Fig. S\ref{fig:reconstruction_amethyst}, S\ref{fig:reconstruction_bunny}, S\ref{fig:reconstruction_boxes}, S\ref{fig:reconstruction_dice}, S\ref{fig:reconstruction_drago}, S\ref{fig:reconstruction_jb}, we illustrate the reconstructed layers using four scalable levels with four images per level along with the original layers. 

\subsection{Comparative Analysis}
\begin{figure*}[!t]
\renewcommand{\figurename}{Figure S}
     \centering
     \begin{subfigure}[b]{0.39\textwidth}
         \centering
         \includegraphics[width=1\textwidth]{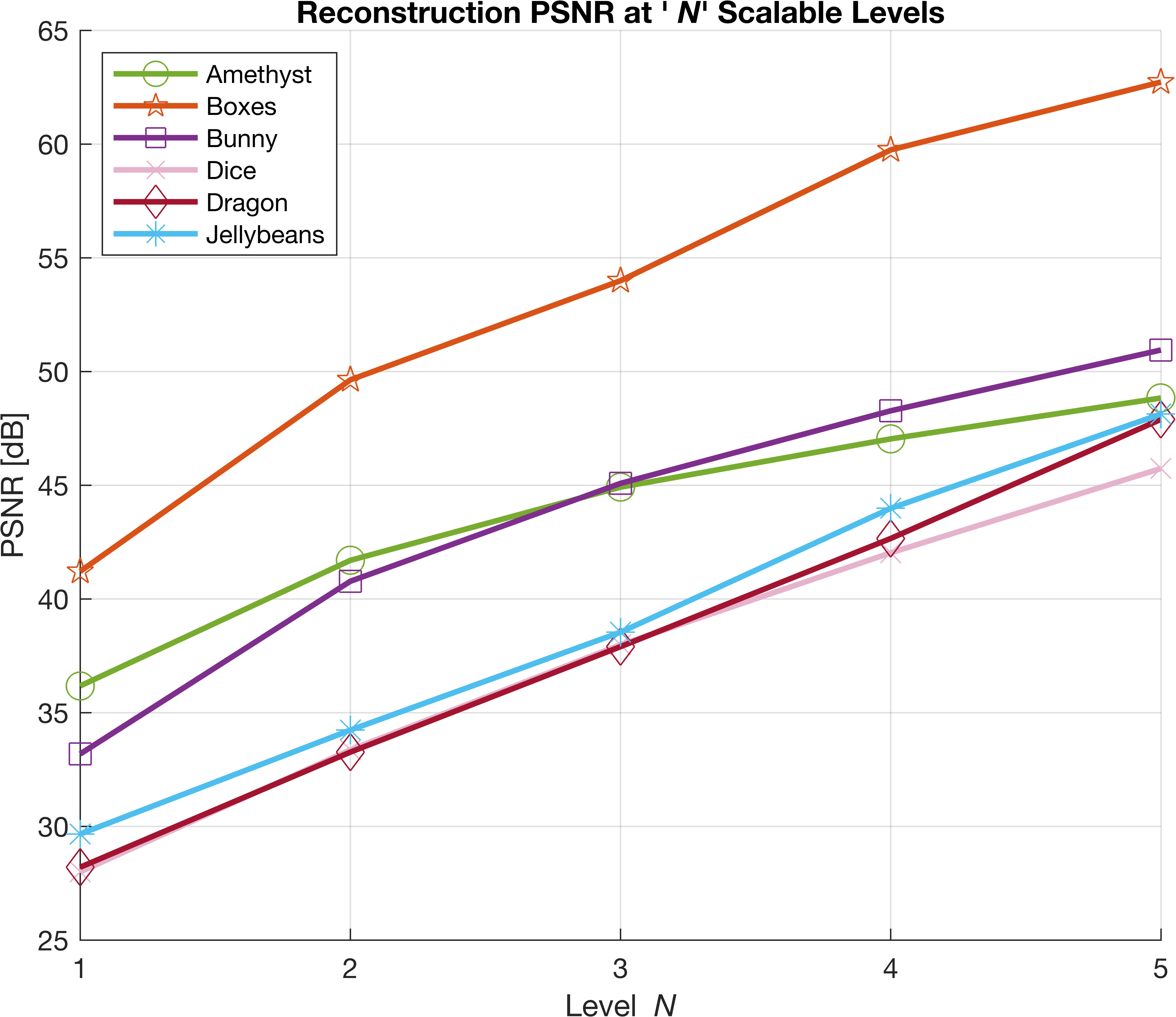}
         \caption{}
     \end{subfigure}
     \begin{subfigure}[b]{0.55\textwidth}
         \centering
         \includegraphics[width=1\textwidth]{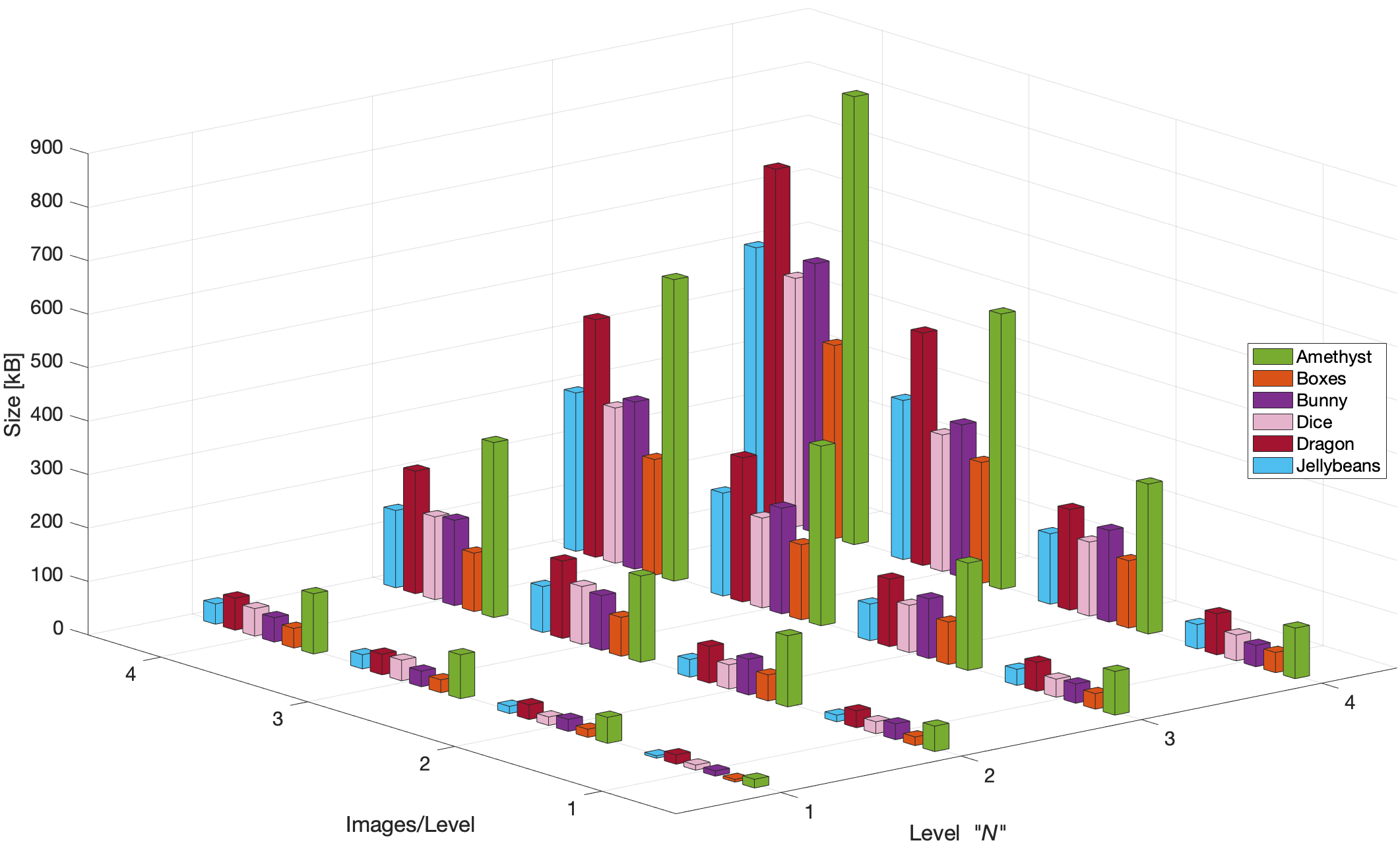}
         \caption{}
     \end{subfigure}
        \caption{Comparison of $(a)$ peak signal-to-noise ratio (PSNR) and $(b)$ sizes between scalable and non-scalable codings with $10$ binary images per level.}
        \label{fig:scalable_non_scalable_psnr_size}
\end{figure*}

Based on our experimental results, we can make some significant conclusions. Firstly, our proposed coding scheme demonstrates remarkable flexibility in producing high-quality reconstructions at higher bitrates. However, the conventional codecs do not support such scenarios. Additionally, it is able to deliver quality on par compared to HEVC even at lower bitrates, which is impressive. To achieve scalability in the pipeline, we implement a multi-level coding approach using the divide-and-conquer method, reducing computational complexity. Furthermore, as illustrated in Fig.~S\ref{fig:scalable_non_scalable_psnr_size}, the decoding accuracy improves progressively as we incorporate more binary images, enhancing the overall efficiency of the system.

\section{Conclusion}
We introduce an efficient representation and scalable coding scheme of a light field for layered light-field displays based on a DBN. Existing light field compression algorithms do not utilise the similarities between the views of a light field to compress them effectively. The proposed scheme exploits the additive layers' spatial, temporal, angular and other non-linear intrinsic redundancies. The proposed hybrid CNN and DBN model generates the optimised transmittance patterns, while learning and preserving only the essential features in a latent code form. Weighted binary images represent the optimised layer patterns in the intermediate stage, where we introduce scalability in the framework. Further encoding with HEVC generates a rate scalable bitstream which supports any desired bitrate (quality) decoding. The decoding performance progressively improves as we use more encoded information in the reconstruction process. As a result, our proposed scheme achieves the goal of covering a wide range of bitrates within a single trained model without compromising the quality of the reconstruction. Experiments with benchmark light field datasets produce competitive results. Since additive layered display is a transparent autostereoscopic display, it can be easily adopted to AR applications. Unlike other light field display, the HOE layers in additive display do not suffer from moir\'{e} effect. We plan to test our scheme with physical light field display hardware in the future. Furthermore, we would like to improve the proposed scheme to deliver 3D content on devices with limited hardware resources.

\bibliographystyle{unsrtnat}
\bibliography{root} 

\begin{thebibliography}{41}
\providecommand{\natexlab}[1]{#1}
\providecommand{\url}[1]{\texttt{#1}}
\expandafter\ifx\csname urlstyle\endcsname\relax
  \providecommand{\doi}[1]{doi: #1}\else
  \providecommand{\doi}{doi: \begingroup \urlstyle{rm}\Url}\fi

\bibitem[Surman and Sun(2014)]{surman2014towards}
Phil Surman and Xiao~Wei Sun.
\newblock Towards the reality of 3d imaging and display.
\newblock In \emph{2014 3DTV-Conference}, pages 1--4, 2014.

\bibitem[Li et~al.(2020)Li, Huang, Alfaro, Supikov, Ratcliff, Grover, and
  Azuma]{li2020light}
Tuotuo Li, Qiong Huang, Santiago Alfaro, Alexey Supikov, Joshua Ratcliff, Ginni
  Grover, and Ronald Azuma.
\newblock Light-field displays: a view-dependent approach.
\newblock In \emph{ACM SIGGRAPH 2020 Emerging Technologies}, pages 1--2. 2020.

\bibitem[Watanabe et~al.(2019)Watanabe, Okaichi, Omura, Kano, Sasaki, and
  Kawakita]{watanabe2019aktina}
Hayato Watanabe, Naoto Okaichi, Takuya Omura, Masanori Kano, Hisayuki Sasaki,
  and Masahiro Kawakita.
\newblock Aktina vision: Full-parallax three-dimensional display with 100
  million light rays.
\newblock \emph{Scientific reports}, 9\penalty0 (1):\penalty0 1--9, 2019.

\bibitem[Ives(1903)]{ives1903parallax}
Frederic~E Ives.
\newblock Parallax stereogram and process of making same., April~14 1903.
\newblock US Patent 725,567.

\bibitem[Isono et~al.(1993)Isono, Yasuda, and
  Sasazawa]{isono1993autostereoscopic}
Haruo Isono, Minoru Yasuda, and Hideaki Sasazawa.
\newblock Autostereoscopic 3-d display using lcd-generated parallax barrier.
\newblock \emph{Electronics and Communications in Japan (Part II:
  Electronics)}, 76\penalty0 (7):\penalty0 77--84, 1993.

\bibitem[Sakamoto and Morii(2006)]{sakamoto2006multi}
Kunio Sakamoto and Tsutomu Morii.
\newblock Multi-view 3d display using parallax barrier combined with polarizer.
\newblock In \emph{Advanced Free-Space Optical Communication Techniques},
  volume 6399, pages 214--221. SPIE, 2006.

\bibitem[Peterka et~al.(2008)Peterka, Kooima, Sandin, Johnson, Leigh, and
  DeFanti]{peterka2008advances}
Tom Peterka, Robert~L Kooima, Daniel~J Sandin, Andrew Johnson, Jason Leigh, and
  Thomas~A DeFanti.
\newblock Advances in the dynallax solid-state dynamic parallax barrier
  autostereoscopic visualization display system.
\newblock \emph{IEEE TCVG}, 14\penalty0 (3):\penalty0 487--499, 2008.

\bibitem[Lippmann(1908)]{lippmann1908epreuves}
Gabriel Lippmann.
\newblock Epreuves reversibles donnant la sensation du relief.
\newblock \emph{J. Phys. Theor. Appl.}, 7\penalty0 (1):\penalty0 821--825,
  1908.

\bibitem[B{\"o}rner(1993)]{borner1993autostereoscopic}
R~B{\"o}rner.
\newblock Autostereoscopic 3d-imaging by front and rear projection and on flat
  panel displays.
\newblock \emph{Displays}, 14\penalty0 (1):\penalty0 39--46, 1993.

\bibitem[McCormick(1995)]{mccormick1995integral}
M~McCormick.
\newblock Integral 3-d imaging for broadcast.
\newblock In \emph{Proc., of the Second International Displays Workshop}, pages
  77--80, 1995.

\bibitem[Wetzstein et~al.(2012)Wetzstein, Lanman, Hirsch, and
  Raskar]{wetzstein2012tensor}
Gordon Wetzstein, Douglas~R Lanman, Matthew~Waggener Hirsch, and Ramesh Raskar.
\newblock Tensor displays: compressive light field synthesis using multilayer
  displays with directional backlighting.
\newblock 2012.

\bibitem[Takahashi et~al.(2018)Takahashi, Kobayashi, and
  Fujii]{takahashi2018focal}
Keita Takahashi, Yuto Kobayashi, and Toshiaki Fujii.
\newblock From focal stack to tensor light-field display.
\newblock \emph{IEEE TIP}, 27\penalty0 (9):\penalty0 4571--4584, 2018.

\bibitem[Maruyama et~al.(2020)Maruyama, Takahashi, and
  Fujii]{maruyama2020comparison}
Keita Maruyama, Keita Takahashi, and Toshiaki Fujii.
\newblock Comparison of layer operations and optimization methods for light
  field display.
\newblock \emph{IEEE Access}, 8:\penalty0 38767--38775, 2020.

\bibitem[Lee et~al.(2016)Lee, Jang, Moon, Cho, and
  Lee]{10.1145/2897824.2925971}
Seungjae Lee, Changwon Jang, Seokil Moon, Jaebum Cho, and Byoungho Lee.
\newblock Additive light field displays: Realization of augmented reality with
  holographic optical elements.
\newblock 35\penalty0 (4), 2016.
\newblock ISSN 0730-0301.

\bibitem[Kramida(2015)]{kramida2015resolving}
Gregory Kramida.
\newblock Resolving the vergence-accommodation conflict in head-mounted
  displays.
\newblock \emph{IEEE TVCG}, 22\penalty0 (7):\penalty0 1912--1931, 2015.

\bibitem[Konrad et~al.(2020)Konrad, Angelopoulos, and
  Wetzstein]{konrad2020gaze}
Robert Konrad, Anastasios Angelopoulos, and Gordon Wetzstein.
\newblock Gaze-contingent ocular parallax rendering for virtual reality.
\newblock \emph{ACM TOG}, 39\penalty0 (2):\penalty0 1--12, 2020.

\bibitem[Sluka et~al.(2021)Sluka, Kvasov, and Kubes]{sluka2021digital}
Tomas Sluka, Alexander Kvasov, and Tomas Kubes.
\newblock Digital light-field.
\newblock 2021.

\bibitem[Liu et~al.(2016)Liu, Wang, Li, Xiong, Wu, and Zeng]{liu2016pseudo}
Dong Liu, Lizhi Wang, Li~Li, Zhiwei Xiong, Feng Wu, and Wenjun Zeng.
\newblock Pseudo-sequence-based light field image compression.
\newblock In \emph{2016 ICMEW}, pages 1--4, 2016.

\bibitem[Li et~al.(2017)Li, Li, Li, Liu, and Li]{li2017pseudo}
Li~Li, Zhu Li, Bin Li, Dong Liu, and Houqiang Li.
\newblock Pseudo-sequence-based 2-d hierarchical coding structure for
  light-field image compression.
\newblock \emph{IEEE Journal of Selected Topics in Signal Processing},
  11\penalty0 (7):\penalty0 1107--1119, 2017.

\bibitem[Sullivan et~al.(2012)Sullivan, Ohm, Han, and
  Wiegand]{sullivan2012overview}
Gary~J Sullivan, Jens-Rainer Ohm, Woo-Jin Han, and Thomas Wiegand.
\newblock Overview of the high efficiency video coding (hevc) standard.
\newblock \emph{IEEE TCSVT}, 22\penalty0 (12):\penalty0 1649--1668, 2012.

\bibitem[Hannuksela et~al.(2015)Hannuksela, Yan, Huang, and
  Li]{hannuksela2015overview}
Miska~M Hannuksela, Ye~Yan, Xuehui Huang, and Houqiang Li.
\newblock Overview of the multiview high efficiency video coding (mv-hevc)
  standard.
\newblock In \emph{2015 ICIP}, pages 2154--2158, 2015.

\bibitem[Senoh et~al.(2018)Senoh, Yamamoto, Tetsutani, and
  Yasuda]{senoh2018efficient}
Takanori Senoh, Kenji Yamamoto, Nobuji Tetsutani, and Hiroshi Yasuda.
\newblock Efficient light field image coding with depth estimation and view
  synthesis.
\newblock In \emph{EUSIPCO}, pages 1840--1844, 2018.

\bibitem[Huang et~al.(2018)Huang, An, Shan, Ma, and Shen]{huang2018view}
Xinpeng Huang, Ping An, Liang Shan, Ran Ma, and Liquan Shen.
\newblock View synthesis for light field coding using depth estimation.
\newblock In \emph{ICME}, pages 1--6. IEEE, 2018.

\bibitem[H{\'e}riard-Dubreuil et~al.(2019)H{\'e}riard-Dubreuil, Viola, and
  Ebrahimi]{heriard2019light}
Baptiste H{\'e}riard-Dubreuil, Irene Viola, and Touradj Ebrahimi.
\newblock Light field compression using translation-assisted view estimation.
\newblock In \emph{Picture Coding Symposium}, pages 1--5, 2019.

\bibitem[Bakir et~al.(2018)Bakir, Hamidouche, D{\'e}forges, Samrouth, and
  Khalil]{bakir2018light}
Nader Bakir, Wassim Hamidouche, Olivier D{\'e}forges, Khouloud Samrouth, and
  Mohamad Khalil.
\newblock Light field image compression based on convolutional neural networks
  and linear approximation.
\newblock In \emph{2018 25th ICIP}, pages 1128--1132, 2018.

\bibitem[Wang et~al.(2019)Wang, Peng, Wang, Han, and Xiang]{wang2019region}
Bing Wang, Qiang Peng, Eric Wang, Kang Han, and Wei Xiang.
\newblock Region-of-interest compression and view synthesis for light field
  video streaming.
\newblock \emph{IEEE Access}, 7:\penalty0 41183--41192, 2019.

\bibitem[Jia et~al.(2018)Jia, Zhang, Wang, Wang, and Ma]{jia2018light}
Chuanmin Jia, Xinfeng Zhang, Shanshe Wang, Shiqi Wang, and Siwei Ma.
\newblock Light field image compression using generative adversarial
  network-based view synthesis.
\newblock \emph{IEEE Journal on Emerging and Selected Topics in Circuits and
  Systems}, 9\penalty0 (1):\penalty0 177--189, 2018.

\bibitem[Liu et~al.(2021)Liu, Huang, Zhan, Ai, Zheng, and Cheng]{liu2021view}
Deyang Liu, Xinpeng Huang, Wenfa Zhan, Liefu Ai, Xin Zheng, and Shulin Cheng.
\newblock View synthesis-based light field image compression using a generative
  adversarial network.
\newblock \emph{Information Sciences}, 545:\penalty0 118--131, 2021.

\bibitem[Vagharshakyan et~al.(2017)Vagharshakyan, Bregovic, and
  Gotchev]{vagharshakyan2017light}
Suren Vagharshakyan, Robert Bregovic, and Atanas Gotchev.
\newblock Light field reconstruction using shearlet transform.
\newblock \emph{IEEE transactions on pattern analysis and machine
  intelligence}, 40\penalty0 (1):\penalty0 133--147, 2017.

\bibitem[Ahmad et~al.(2020)Ahmad, Vagharshakyan, Sj{\"o}str{\"o}m, Gotchev,
  Bregovic, and Olsson]{ahmad2020shearlet}
Waqas Ahmad, Suren Vagharshakyan, M{\aa}rten Sj{\"o}str{\"o}m, Atanas Gotchev,
  Robert Bregovic, and Roger Olsson.
\newblock Shearlet transform-based light field compression under low bitrates.
\newblock \emph{IEEE TIP}, 29:\penalty0 4269--4280, 2020.

\bibitem[Chen et~al.(2020)Chen, An, Huang, Yang, Liu, and Wu]{chen2020light}
Yilei Chen, Ping An, Xinpeng Huang, Chao Yang, Deyang Liu, and Qiang Wu.
\newblock Light field compression using global multiplane representation and
  two-step prediction.
\newblock \emph{IEEE Signal Processing Letters}, 27:\penalty0 1135--1139, 2020.

\bibitem[Teh and Hinton(2000)]{teh2000rate}
Yee~Whye Teh and Geoffrey~E Hinton.
\newblock Rate-coded restricted boltzmann machines for face recognition.
\newblock \emph{Advances in neural information processing systems}, 13, 2000.

\bibitem[Hinton and Salakhutdinov(2006)]{hinton2006reducing}
Geoffrey~E Hinton and Ruslan~R Salakhutdinov.
\newblock Reducing the dimensionality of data with neural networks.
\newblock \emph{science}, 313\penalty0 (5786):\penalty0 504--507, 2006.

\bibitem[Komatsu et~al.(2018)Komatsu, Takahashi, and Fujii]{8451812}
Koji Komatsu, Keita Takahashi, and Toshiaki Fujii.
\newblock Scalable light field coding using weighted binary images.
\newblock In \emph{IEEE ICIP}, pages 903--907, 2018.

\bibitem[Toth(2000)]{toth2000optimization}
Paolo Toth.
\newblock Optimization engineering techniques for the exact solution of np-hard
  combinatorial optimization problems.
\newblock \emph{European journal of operational research}, 125\penalty0
  (2):\penalty0 222--238, 2000.

\bibitem[Shang et~al.(2014)Shang, Yang, Huang, and Lyu]{shang2014data}
Chao Shang, Fan Yang, Dexian Huang, and Wenxiang Lyu.
\newblock Data-driven soft sensor development based on deep learning technique.
\newblock \emph{Journal of Process Control}, 24\penalty0 (3):\penalty0
  223--233, 2014.

\bibitem[Hinton(2002)]{hinton2002training}
Geoffrey~E Hinton.
\newblock Training products of experts by minimizing contrastive divergence.
\newblock \emph{Neural computation}, 14\penalty0 (8):\penalty0 1771--1800,
  2002.

\bibitem[Vaish and Adams(2008)]{vaish2008new}
Vaibhav Vaish and Andrew Adams.
\newblock The (new) stanford light field archive.
\newblock \emph{Computer Graphics Laboratory, Stanford University}, 6\penalty0
  (7):\penalty0 3, 2008.

\bibitem[Honauer et~al.(2017)Honauer, Johannsen, Kondermann, and
  Goldluecke]{honauer2017dataset}
Katrin Honauer, Ole Johannsen, Daniel Kondermann, and Bastian Goldluecke.
\newblock A dataset and evaluation methodology for depth estimation on 4d light
  fields.
\newblock In \emph{Computer Vision--ACCV}, pages 19--34, 2017.

\bibitem[Marwah et~al.(2013)Marwah, Wetzstein, Bando, and
  Raskar]{marwah2013compressive}
Kshitij Marwah, Gordon Wetzstein, Yosuke Bando, and Ramesh Raskar.
\newblock Compressive light field photography using overcomplete dictionaries
  and optimized projections.
\newblock \emph{ACM TOG}, 32\penalty0 (4):\penalty0 1--12, 2013.

\bibitem[Bjontegaard(2001)]{bjontegaard2001calculation}
Gisle Bjontegaard.
\newblock Calculation of average psnr differences between rd-curves.
\newblock \emph{VCEG-M33}, 2001.

\end{thebibliography}

\section{Supplementary Material:}

\begin{table*}[!h]
    \caption{\footnotesize The percentage rate savings based on Bjontegaard metric for the proposed compression scheme with respect to HEVC (HM 17.0) on \textit{Amethyst}, \textit{Boxes}, \textit{Bunny}, \textit{Dice}, \textit{DragonAndBunnies} and \textit{Jellybeans} light fields.}
    \centering
    \resizebox{\textwidth}{!}{
    \begin{tabular}{|c|cc|cc|cc|cc|cc|cc|cc|cc|}
    \hline
          & \multicolumn{8}{c|}{\textbf{Amethyst}}                        & \multicolumn{8}{c|}{\textbf{Boxes}} \\
    \hline
          & \multicolumn{2}{c|}{\textbf{Non-Scalable N = 1}} & \multicolumn{2}{c|}{\textbf{Scalable N = 2}} & \multicolumn{2}{c|}{\textbf{Scalable N = 3}} & \multicolumn{2}{c|}{\textbf{Scalable N = 4}} & \multicolumn{2}{c|}{\textbf{Non-Scalable N = 1}} & \multicolumn{2}{c|}{\textbf{Scalable N = 2}} & \multicolumn{2}{c|}{\textbf{Scalable N = 3}} & \multicolumn{2}{c|}{\textbf{Scalable N = 4}} \\
    \hline
    \textbf{QP} & \multicolumn{1}{c|}{\textbf{BD-Rate}} & \textbf{BD-PSNR} & \multicolumn{1}{c|}{\textbf{BD-Rate}} & \textbf{BD-PSNR} & \multicolumn{1}{c|}{\textbf{BD-Rate}} & \textbf{BD-PSNR} & \multicolumn{1}{c|}{\textbf{BD-Rate}} & \textbf{BD-PSNR} & \multicolumn{1}{c|}{\textbf{BD-Rate}} & \textbf{BD-PSNR} & \multicolumn{1}{c|}{\textbf{BD-Rate}} & \textbf{BD-PSNR} & \multicolumn{1}{c|}{\textbf{BD-Rate}} & \textbf{BD-PSNR} & \multicolumn{1}{c|}{\textbf{BD-Rate}} & \textbf{BD-PSNR} \\
    \hline
    \textbf{0} & -98.8690 & 57.6900 & -55.7210 & -8.0804 & -61.2000 & -7.4733 & -24.1800 & -15.2940 & -41.8959 & -11.8468 & 198.1354 & -33.6725 & 169.2788 & -32.8233 & 225.0941 & -35.1330 \\
    \textbf{4} & -98.6660 & 47.7800 & -64.4650 & -13.8590 & -70.9230 & -13.2360 & -46.5880 & -20.7910 & -60.0660 & -16.2634 & 73.3905 & -37.8468 & 57.3914 & -36.8934 & 78.3453 & -39.2278 \\
    \textbf{8} & -95.3180 & 23.4260 & 31.3520 & -32.7900 & 17.1520 & -32.3530 & 116.2800 & -39.8830 & 25.4220 & -34.0146 & 526.7320 & -55.7807 & 483.1239 & -54.7671 & 588.0735 & -57.2147 \\
    \textbf{12} & -86.5830 & 12.1360 & 267.0900 & -43.1290 & 232.8400 & -43.0600 & 528.8800 & -50.7920 & 209.7793 & -46.9153 & 1625.4085 & -69.7935 & 1520.5142 & -68.4530 & 1875.0145 & -70.9254 \\
    \textbf{16} & -66.8830 & 4.0749 & 1095.4000 & -51.0790 & 1053.5000 & -51.6410 & 2135.6000 & -59.2240 & 863.3522 & -58.7275 & 5477.7887 & -82.8921 & 5028.0985 & -81.0795 & 6294.1877 & -83.2878 \\
    \textbf{20} & -30.0260 & 0.4346 & 2155.4000 & -53.1150 & 2173.8000 & -53.5410 & 4320.4000 & -60.7000 & 1883.4621 & -66.4779 & 12072.7062 & -91.7540 & 10746.6098 & -88.9064 & 13314.8035 & -90.9410 \\
    \textbf{24} & -0.8148 & 2.5083 & 2152.0000 & -46.1940 & 1964.9000 & -42.6370 & 3591.6000 & -49.1290 & 1974.4314 & -59.1979 & 11442.4152 & -81.5360 & 8855.1567 & -75.0587 & 9672.0447 & -73.5113 \\
    \textbf{28} & 27.8410 & 5.0623 & 1396.0000 & -38.8420 & 1019.7000 & -29.7390 & 1687.2000 & -35.8920 & 937.6225 & -43.6612 & 3182.6067 & -56.3625 & 2032.0911 & -45.8879 & 1640.0668 & -37.7108 \\
    \textbf{32} & 66.1340 & 4.5937 & 890.8700 & -34.2510 & 570.3500 & -21.8320 & 800.0600 & -27.5280 & 356.6148 & -26.4650 & 587.8523 & -25.4563 & 384.0154 & -16.0203 & 235.1648 & -2.1866 \\
    \textbf{36} & 65.4110 & 3.0468 & 512.5400 & -25.5620 & 385.7100 & -12.5280 & 472.1000 & -17.6450 & 223.1411 & -27.8435 & 119.1653 & 2.3155 & 65.1886 & 12.7977 & 15.0704 & 36.7899 \\
    \textbf{40} & 124.9200 & 6.2937 & 288.7800 & -20.3660 & 259.5800 & -11.6720 & 289.1500 & -16.0210 & 75.2486 & 0.0205 & 39.3248 & 11.2568 & -25.5882 & 74.0820 & -28.8742 & 15.3309 \\
    \textbf{44} & 108.3400 & -5.2522 & 219.9400 & -30.2000 & 200.1300 & -21.2540 & 227.6000 & -27.0140 & 72.1109 & -14.5280 & -7.0705 & 9.7634 & -24.0337 & 13.7988 & -34.9223 & 16.8794 \\
    \textbf{48} & 30.8840 & 13.8740 & 6.6178 & 15.8100 & 23.6090 & 14.4030 & 17.8290 & 14.7260 & -22.1333 & -2.6566 & -28.4868 & -1.3393 & -28.3266 & -1.3673 & -29.5321 & -1.1023 \\
    \textbf{51} & 45.1830 & -10.1580 & 22.7340 & -8.8224 & 39.6500 & -9.9032 & 33.6690 & -9.5537 & -9.7703 & -1.1806 & -16.3486 & 0.2101 & -16.2866 & 0.2038 & -16.5424 & 0.2631 \\
    \hline
          & \multicolumn{8}{|c|}{\textbf{Bunny}}                           & \multicolumn{8}{c|}{\textbf{Dice}} \\
    \hline
          & \multicolumn{2}{c|}{\textbf{Non-Scalable N = 1}} & \multicolumn{2}{c|}{\textbf{Scalable N = 2}} & \multicolumn{2}{c|}{\textbf{Scalable N = 3}} & \multicolumn{2}{c|}{\textbf{Scalable N = 4}} & \multicolumn{2}{c|}{\textbf{Non-Scalable N = 1}} & \multicolumn{2}{c|}{\textbf{Scalable N = 2}} & \multicolumn{2}{c|}{\textbf{Scalable N = 3}} & \multicolumn{2}{c|}{\textbf{Scalable N = 4}} \\
    \hline
    \textbf{QP} & \multicolumn{1}{c|}{\textbf{BD-Rate}} & \textbf{BD-PSNR} & \multicolumn{1}{c|}{\textbf{BD-Rate}} & \textbf{BD-PSNR} & \multicolumn{1}{c|}{\textbf{BD-Rate}} & \textbf{BD-PSNR} & \multicolumn{1}{c|}{\textbf{BD-Rate}} & \textbf{BD-PSNR} & \multicolumn{1}{c|}{\textbf{BD-Rate}} & \textbf{BD-PSNR} & \multicolumn{1}{c|}{\textbf{BD-Rate}} & \textbf{BD-PSNR} & \multicolumn{1}{c|}{\textbf{BD-Rate}} & \textbf{BD-PSNR} & \multicolumn{1}{c|}{\textbf{BD-Rate}} & \textbf{BD-PSNR} \\
    \hline
    \textbf{0} & -35.8541 & -17.0695 & 73.0508 & -32.3806 & 160.6051 & -39.3341 & 624.6934 & -52.8248 & -66.4276 & -2.2326 & 40.6093 & -22.5932 & 130.0192 & -28.8897 & 170.5627 & -30.7740 \\
    \textbf{4} & -41.8936 & -34.0459 & 40.0770 & -49.8663 & 86.2890 & -57.2647 & 371.3065 & -72.0286 & -75.9316 & -6.9096 & -15.6806 & -26.5816 & 22.7096 & -32.4988 & 43.2013 & -34.3269 \\
    \textbf{8} & 175.0928 & -54.4521 & 596.2660 & -71.5034 & 968.0501 & -79.8450 & 2884.9865 & -96.0686 & -33.4818 & -17.8915 & 158.3820 & -36.8031 & 312.4884 & -42.7621 & 399.1539 & -44.3851 \\
    \textbf{12} & 581.4316 & -69.2273 & 1738.6878 & -88.2156 & 3027.9593 & -97.0145 & 8733.8350 & -115.7014 & 29.9154 & -25.3225 & 422.0706 & -44.1848 & 755.3686 & -50.1750 & 956.5430 & -51.6672 \\
    \textbf{16} & 1744.7020 & -81.1327 & 5404.1429 & -101.9941 & 9187.7433 & -111.9698 & 29726.3101 & -132.4633 & 227.5596 & -31.2497 & 1279.9479 & -49.7174 & 2232.8511 & -55.6054 & 2712.3778 & -56.8894 \\
    \textbf{20} & 3007.3424 & -84.3790 & 9496.2841 & -105.7178 & 17104.4903 & -116.9110 & 59241.9373 & -139.1975 & 440.0170 & -31.5654 & 2349.4750 & -49.3073 & 4050.7552 & -54.6757 & 5005.8425 & -55.4995 \\
    \textbf{24} & 2406.6269 & -69.8786 & 6191.8533 & -83.3898 & 10202.4810 & -89.1761 & 33377.0580 & -107.5293 & 472.3484 & -24.1805 & 2008.4562 & -37.2643 & 3373.7957 & -42.3617 & 3824.9943 & -41.7005 \\
    \textbf{28} & 1181.4572 & -53.6379 & 1707.3778 & -52.8878 & 2611.9269 & -58.3495 & 4573.4586 & -60.7393 & 296.4435 & -13.8790 & 878.9704 & -21.6984 & 1222.9146 & -23.3816 & 1264.8145 & -21.9235 \\
    \textbf{32} & 655.5291 & -45.8655 & 603.1516 & -35.9773 & 1128.7717 & -50.2445 & 1084.3950 & -37.4922 & 144.7853 & -4.2159 & 317.2983 & -6.5884 & 349.4932 & -5.1907 & 326.3524 & -2.3271 \\
    \textbf{36} & 317.9121 & -28.6696 & 186.1383 & -12.7256 & 454.1258 & -34.3873 & 282.1265 & -15.3242 & 45.9910 & 6.8538 & 105.9174 & 4.2912 & 106.3284 & 11.0384 & 100.4304 & 8.1394 \\
    \textbf{40} & 143.5400 & -18.2187 & 45.7421 & 7.3104 & 209.0109 & -27.0753 & 108.9672 & -6.3548 & -22.3509 & 29.1828 & 12.5711 & 25.6253 & 10.0207 & 27.5119 & -11.0958 & 33.5574 \\
    \textbf{44} & 125.4312 & -42.8964 & 36.2615 & -9.3876 & 158.8552 & -49.7373 & 38.9808 & -2.4908 & -62.1477 & 56.4491 & -25.8455 & 28.2568 & -24.2835 & 28.0635 & -41.4466 & 37.4671 \\
    \textbf{48} & -4.4405 & -2.4688 & -10.8485 & -1.1827 & -3.5751 & -2.5988 & -6.8336 & -1.9784 & -82.2196 & 22.2027 & -75.5257 & 20.0590 & -76.0856 & 20.2116 & -78.8015 & 21.0589 \\
    \textbf{51} & 7.4971 & -1.0392 & 2.7308 & -0.0387 & 8.3584 & -1.1732 & 5.3396 & -0.5691 & -81.5310 & 22.0370 & -75.0849 & 19.9392 & -75.6145 & 20.0897 & -78.4855 & 21.0813 \\
    \hline
          & \multicolumn{8}{c|}{\textbf{DragonAndBunnies}}                        & \multicolumn{8}{c|}{\textbf{Jellybeans}} \\
    \hline
          & \multicolumn{2}{c|}{\textbf{Non-Scalable N = 1}} & \multicolumn{2}{c|}{\textbf{Scalable N = 2}} & \multicolumn{2}{c|}{\textbf{Scalable N = 3}} & \multicolumn{2}{c|}{\textbf{Scalable N = 4}} & \multicolumn{2}{c|}{\textbf{Non-Scalable N = 1}} & \multicolumn{2}{c|}{\textbf{Scalable N = 2}} & \multicolumn{2}{c|}{\textbf{Scalable N = 3}} & \multicolumn{2}{c|}{\textbf{Scalable N = 4}} \\
    \hline
    \textbf{QP} & \multicolumn{1}{c|}{\textbf{BD-Rate}} & \textbf{BD-PSNR} & \multicolumn{1}{c|}{\textbf{BD-Rate}} & \textbf{BD-PSNR} & \multicolumn{1}{c|}{\textbf{BD-Rate}} & \textbf{BD-PSNR} & \multicolumn{1}{c|}{\textbf{BD-Rate}} & \textbf{BD-PSNR} & \multicolumn{1}{c|}{\textbf{BD-Rate}} & \textbf{BD-PSNR} & \multicolumn{1}{c|}{\textbf{BD-Rate}} & \textbf{BD-PSNR} & \multicolumn{1}{c|}{\textbf{BD-Rate}} & \textbf{BD-PSNR} & \multicolumn{1}{c|}{\textbf{BD-Rate}} & \textbf{BD-PSNR} \\
    \hline
    \textbf{0} & -85.5156 & 8.3380 & 32.0122 & -20.2387 & 72.9238 & -23.5259 & 66.2699 & -22.0434 & -99.5234 & 77.9058 & -76.7885 & -1.9367 & -54.4376 & -11.1967 & 125.1849 & -29.2147 \\
    \textbf{4} & -90.2718 & 4.0256 & -33.2695 & -23.2836 & -12.0985 & -26.6480 & -15.2971 & -25.0874 & -99.3836 & 55.7624 & -82.6403 & -15.0227 & -69.6617 & -23.9664 & 23.3035 & -42.0870 \\
    \textbf{8} & -72.5248 & -5.6574 & 131.7141 & -31.8302 & 212.8679 & -34.8821 & 196.4963 & -33.1752 & -96.7078 & 33.6512 & -1.4889 & -32.4229 & 88.8461 & -41.9025 & 773.3147 & -60.4853 \\
    \textbf{12} & -48.0885 & -12.1782 & 378.6404 & -37.9446 & 544.6392 & -40.9934 & 514.1859 & -38.8798 & -91.5477 & 22.1932 & 177.6594 & -42.7123 & 455.2631 & -52.9002 & 2623.2910 & -72.1530 \\
    \textbf{16} & 26.6078 & -17.6267 & 1153.7395 & -42.8081 & 1590.5743 & -45.7413 & 1473.7028 & -43.0534 & -77.3925 & 13.0968 & 831.6044 & -51.5420 & 1925.2471 & -62.6116 & 10388.1426 & -82.3759 \\
    \textbf{20} & 110.3281 & -18.3464 & 2105.4681 & -41.5987 & 2906.9074 & -44.2775 & 2430.3875 & -40.7478 & -48.8109 & 11.0514 & 1693.8994 & -50.8611 & 4117.8794 & -62.3073 & 23403.4191 & -81.2988 \\
    \textbf{24} & 123.8702 & -10.5361 & 1665.4596 & -31.4250 & 2048.6159 & -30.8901 & 1559.1508 & -27.8501 & -38.5762 & 21.9889 & 1539.6563 & -38.8265 & 3515.4397 & -47.9189 & 18169.2832 & -65.0034 \\
    \textbf{28} & 56.7639 & -0.7447 & 655.9266 & -16.4986 & 592.6371 & -11.8968 & 451.1796 & -8.8809 & -54.2492 & 49.7829 & 833.6331 & -23.1061 & 1381.1249 & -27.2223 & 4955.1499 & -39.5767 \\
    \textbf{32} & 2.3774 & 9.0746 & 172.9573 & 0.5472 & 121.8285 & 6.8847 & 86.1353 & 9.6809 & -59.9947 & 66.5945 & 323.5571 & -6.5438 & 423.1734 & -7.6443 & 677.8773 & -6.1105 \\
    \textbf{36} & -41.3905 & 29.7503 & 7.3674 & 18.8344 & -3.6047 & 22.8576 & -22.2246 & 29.4171 & -65.6308 & 95.8310 & 140.9084 & 12.6884 & 160.4089 & 12.6442 & 6.4096 & 57.1260 \\
    \textbf{40} & -68.2151 & 57.5858 & -51.7591 & 46.3843 & -52.5853 & 47.4873 & -60.6794 & 58.4989 & -58.1097 & 87.6541 & 37.2624 & 30.8651 & 26.7779 & 35.6626 & -56.4936 & 138.6558 \\
    \textbf{44} & -78.7734 & 97.5241 & -68.6222 & 62.3341 & -66.4932 & 57.4959 & -75.4121 & 95.2259 & -76.5525 & 22.3631 & 3.0126 & 28.2061 & 2.2206 & 30.3434 & -70.5624 & 20.7238 \\
    \textbf{48} & -85.9083 & 23.2501 & -82.5366 & 21.8436 & -81.8672 & 21.5962 & -83.4843 & 22.2759 & -73.0002 & 21.4228 & -56.0492 & 17.6430 & -55.6653 & 17.5842 & -67.0470 & 19.8675 \\
    \textbf{51} & -83.3053 & 22.6011 & -80.4031 & 21.4844 & -79.9767 & 21.3327 & -80.8549 & 21.7928 & -70.1418 & 20.4150 & -53.9243 & 16.8270 & -53.4734 & 16.7459 & -64.5351 & 18.9824 \\
    \hline
    \end{tabular}%
    }
  \label{tab:bdratepsnr}%
\end{table*}%

\end{document}